\makeatletter\def\graphicscache@inhibit{true}\makeatother
\documentclass{llncs}
\usepackage{makeidx}  %
\usepackage[utf8]{inputenc}
\usepackage{lmodern}
\usepackage[T1]{fontenc}
\usepackage{microtype}
\usepackage{textcomp}
\usepackage{graphicx} %

\usepackage{pgfplots}
\pgfplotsset{compat=newest}
\usepackage{subfig}
\usepackage{multirow}
\usepackage{tabularx}
\usepackage{threeparttable}
\usepackage{booktabs}
\usepackage{array}
\usepackage{float}
\usepackage{graphbox}
\usepackage{tikz,tikz-3dplot}
\usetikzlibrary{fit,arrows,arrows.meta,automata,backgrounds,calc,chains,%
decorations.markings,decorations.pathreplacing,decorations.pathmorphing,%
matrix,positioning,shapes,shapes.geometric,shapes.symbols,spy,trees,tikzmark}
\usepackage{balance}
\usepackage[binary-units=true,product-units=single,per-mode=symbol,range-units=single,range-phrase=\,--\,,detect-all]{siunitx}
\DeclareSIUnit\pixel{px}
\usepackage{amsmath} %
\usepackage{amssymb}  %
\usepackage{bm}
\usepackage{acronym}
\usepackage{tablefootnote}

\definecolor{bg_color}{RGB}{95,95,95}

\let\vec\bm

\newcommand{\reffig}[1]{Fig.~\ref{#1}}
\newcommand{\reftab}[1]{Tab.~\ref{#1}}
\newcommand{\refsec}[1]{Sec.~\ref{#1}}

\newcommand{\etal}{et al.~}

\newcommand{\wrt}{~w.r.t.~}

\newcommand{\eg}{e.g.,\ }
\newcommand{\cf}{cf.\ }

\usepackage{adjustbox}
\newcolumntype{R}[2]{%
    >{\adjustbox{angle=#1,lap=\width-(#2)}\bgroup}%
    l%
    <{\egroup}%
}
\newcolumntype{L}[1]{>{\raggedright\let\newline\\\arraybackslash\hspace{0pt}}m{#1}}

\begin{document}
\frontmatter          %
\pagestyle{headings}  %
\addtocmark{3D Semantic Scene Perception using Distributed Smart Edge Sensors} %
\mainmatter              %
\title{3D Semantic Scene Perception using Distributed Smart Edge Sensors}
\titlerunning{3D Semantic Scene Perception using Distributed Smart Edge Sensors}  %
\author{Simon Bultmann \and Sven Behnke}
\authorrunning{S. Bultmann and S. Behnke} %
\tocauthor{Simon Bultmann and Sven Behnke}
\institute{Institute for Computer Science VI, Autonomous Intelligent Systems,\\
University of Bonn, Friedrich-Hirzebruch-Allee 8, 53115 Bonn, Germany\\
\email{bultmann@ais.uni-bonn.de}, \texttt{https://www.ais.uni-bonn.de}
}

\maketitle              %

\begin{tikzpicture}[remember picture,overlay]
\node[anchor=north west,align=left,font=\sffamily,yshift=-0.2cm,xshift=0.2cm] at (current page.north west) {%
  17th International Conference on Intelligent Autonomous Systems (IAS), Zagreb, Croatia, June 2022.
};
\end{tikzpicture}%
\vspace{-0.2cm}%

\begin{abstract}
We present a system for 3D semantic scene perception consisting of a network of distributed smart edge sensors. The sensor nodes are based on an embedded CNN inference accelerator and RGB-D and thermal cameras.
Efficient vision CNN models for object detection, semantic segmentation, and human pose estimation run on-device in real time.
2D human keypoint estimations, augmented with the RGB-D depth estimate, as well as semantically annotated point clouds are streamed from the sensors to a central backend, where multiple viewpoints are fused into an allocentric 3D semantic scene model.
As the image interpretation is computed locally, only semantic information is sent over the network. The raw images remain on the sensor boards, significantly reducing the required bandwidth, and mitigating privacy risks for the observed persons.

We evaluate the proposed system in challenging real-world multi-person scenes in our lab.
The proposed perception system provides a complete scene view containing semantically annotated 3D geometry and estimates 3D poses of multiple persons in real time.

\end{abstract}

\section{Introduction}
\label{sec:Introduction}
Accurate semantic perception of 3D scene geometry and persons is challenging and an important prerequisite for many robotic tasks, such as safe and anticipative robot movement in the vicinity of people as well as human-robot interaction. 
In this work, we propose a system for 3D semantic scene perception consisting of a network of distributed smart edge sensors. It provides a complete scene view containing semantically annotated 3D geometry and estimates 3D poses of multiple persons in real time.
\begin{figure}[t]
	\centering
		\begin{tikzpicture}
			\node[inner sep=0,anchor=north west] (image1) at (0, 0) {\includegraphics[height=4.cm,trim= 2500 900 20 550, clip]{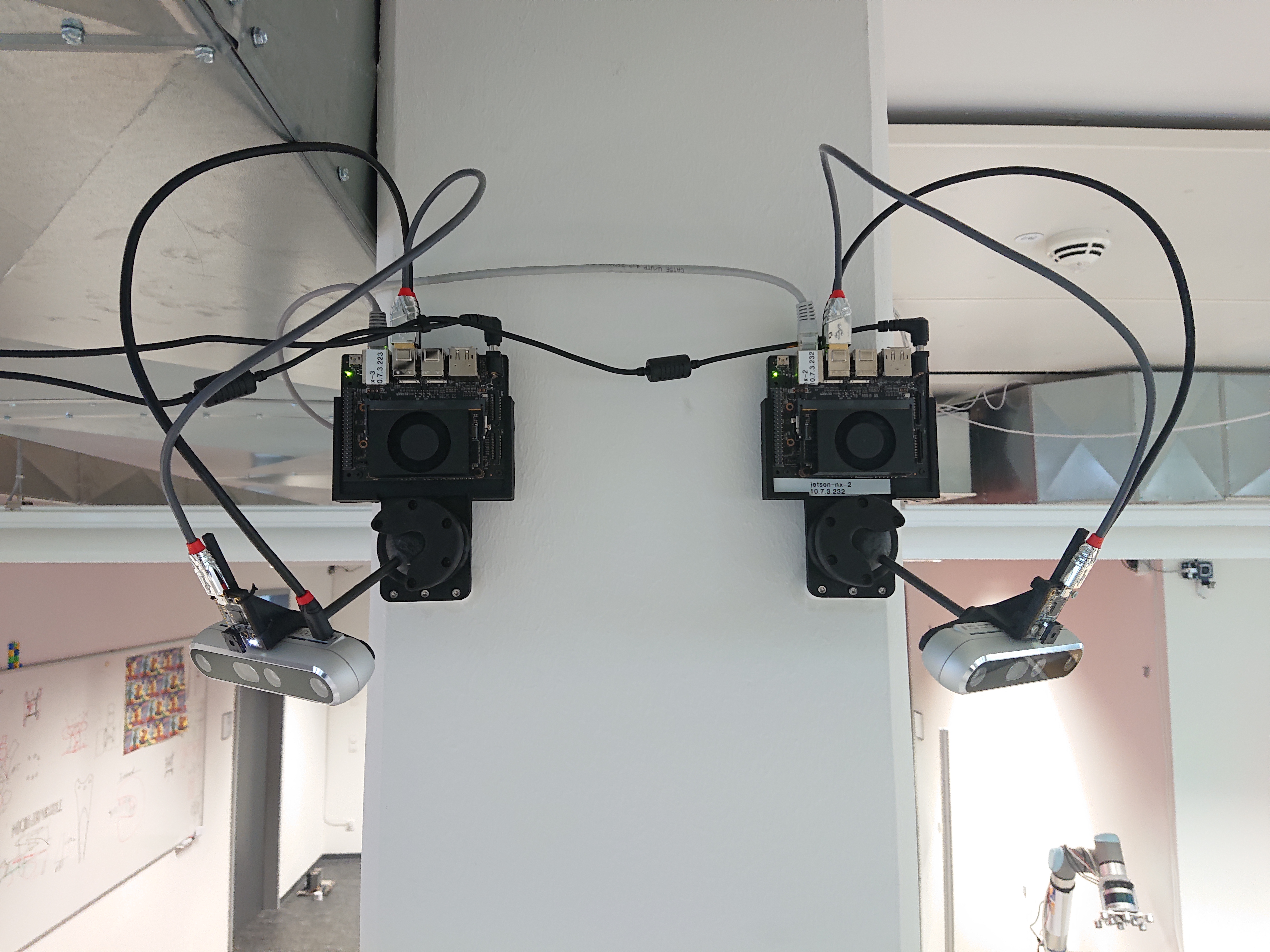}};
			\node[inner sep=0,anchor=north west,xshift=0.1cm] (image2) at (image1.north east) {\includegraphics[height=4.cm]{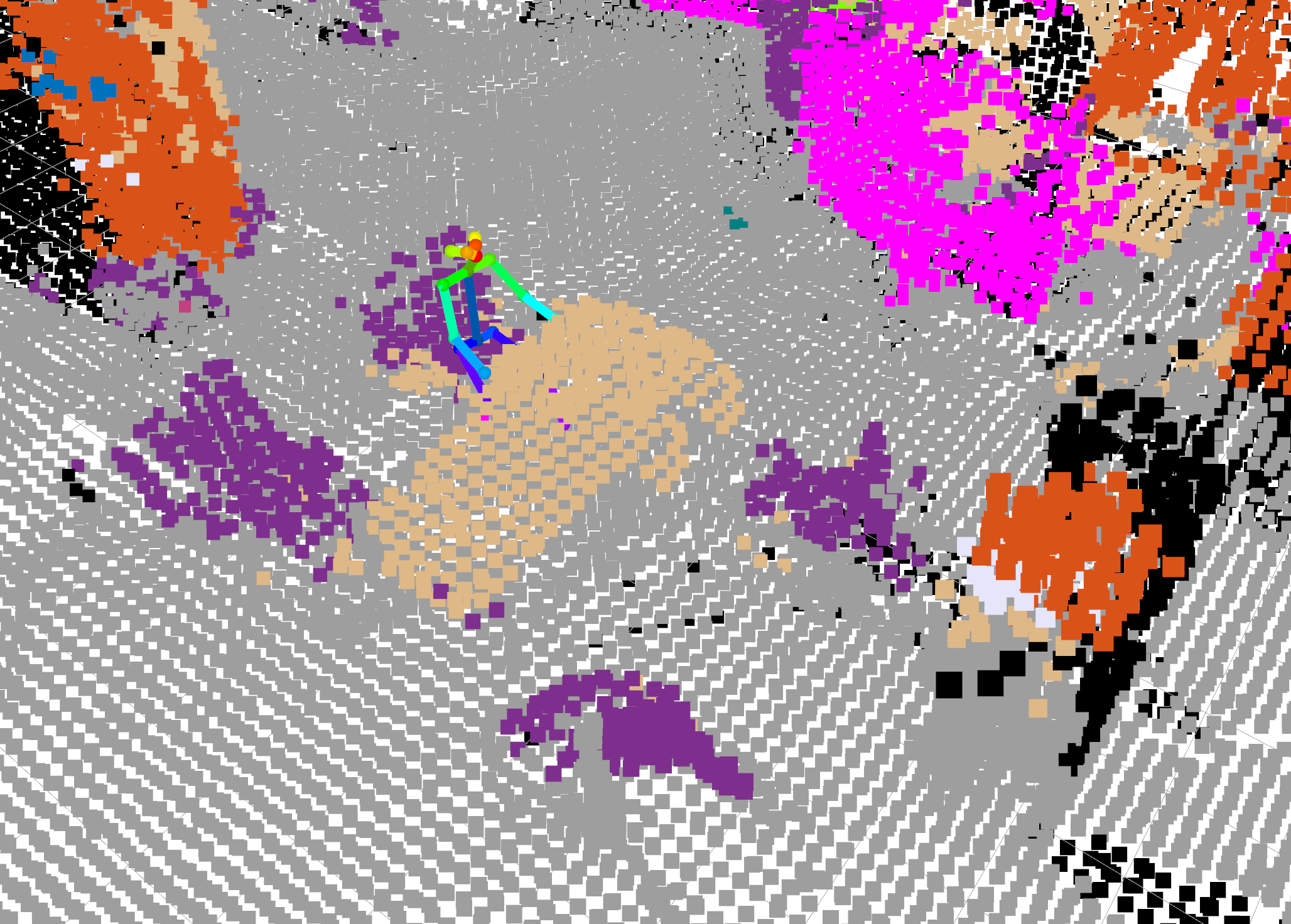}};
			\node[inner sep=0,anchor=north west,xshift=0.1cm] (image3) at (image2.north east) {\includegraphics[height=4.cm]{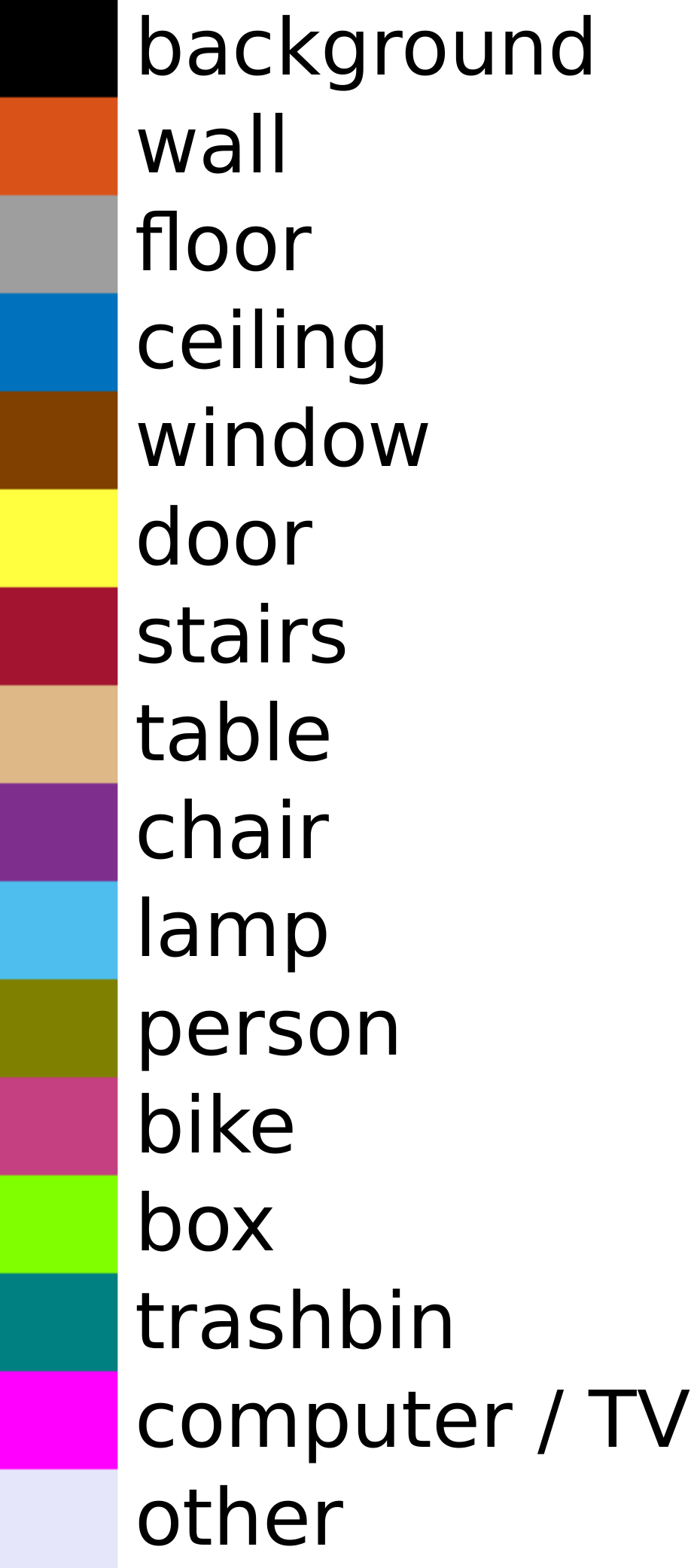}};
			\node[inner sep=0,anchor=north west,yshift=-0.1cm] (image4) at (image1.south west){\includegraphics[height=2.45cm]{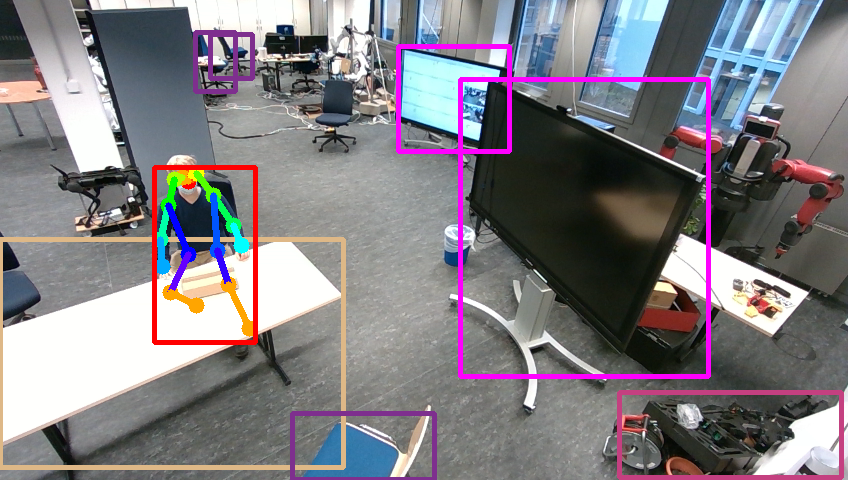}};
			\node[inner sep=0,anchor=north west,xshift=0.1cm] (image5) at (image4.north east) {\includegraphics[height=2.45cm]{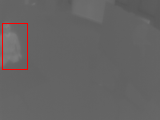}};
			\node[inner sep=0,anchor=south west,xshift=0.1cm] (image6) at (image5.south east) {\includegraphics[height=2.45cm]{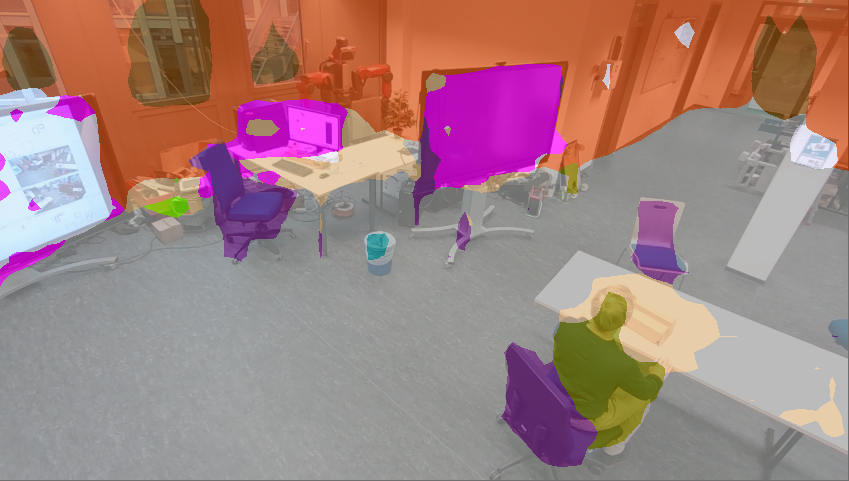}};
			
			\node[label,scale=.75, anchor=south west, rectangle, fill=white, align=center, font=\scriptsize\sffamily] (n_0) at (image1.south west) {(a)};
			\node[label,scale=.75, anchor=south west, rectangle, fill=white, align=center, font=\scriptsize\sffamily] (n_1) at (image2.south west) {(b)};
			\node[label,scale=.75, anchor=south west, rectangle, fill=white, align=center, font=\scriptsize\sffamily] (n_3) at (image4.south west) {(c)};
			\node[label,scale=.75, anchor=south west, rectangle, fill=white, align=center, font=\scriptsize\sffamily] (n_4) at (image5.south west) {(d)};
			\node[label,scale=.75, anchor=south west, rectangle, fill=white, align=center, font=\scriptsize\sffamily] (n_5) at (image6.south west) {(e)};
		\end{tikzpicture}
	\caption{Semantic perception with distributed smart edge sensors: (a) developed sensor node, (b) 3D semantic scene model with 3D human skeleton, (c) RGB and (d) thermal detections, (e) semantic segmentation. Person detections in red and skeleton keypoints colored by joint index. Occluded joints are marked in orange. CNN inference runs online on distributed sensors and semantic information is aggregated into an allocentric 3D scene model on the backend including 3D geometry (e.g., furniture, walls, floor) and 3D human pose.}
	\label{fig:teaser}
	\vspace{-1em}
\end{figure}
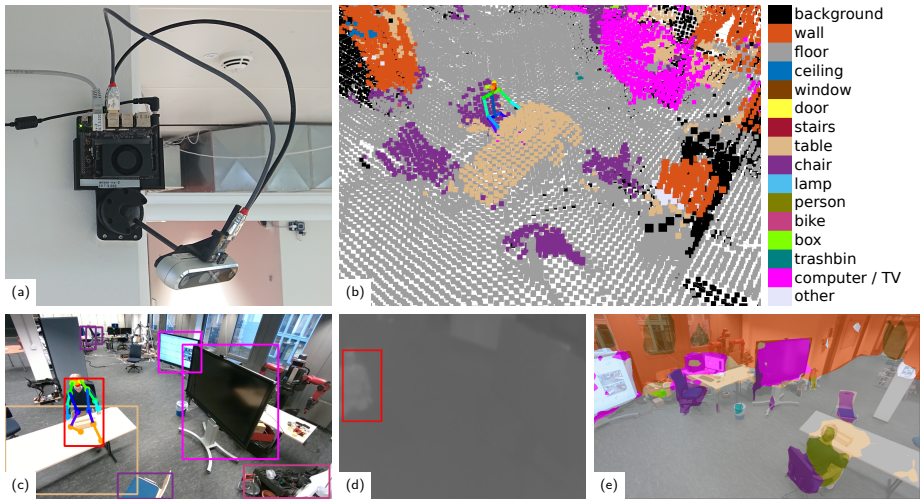

We build upon our previous work on real-time 3D human pose estimation using semantic feedback to smart edge sensors~\cite{Bultmann_RSS_2021}.
While this existing pipeline is able to track poses of multiple persons in real time, it lacks modeling of other aspects of the scene, i.e. 3D geometry, object detections, and surface categorization. Semantically annotated 3D geometry, however, is required to explain and predict interactions between persons and objects in the scene, and to handle occlusions. Temporal aggregation and fusion of semantic point clouds from multiple sensor perspectives further leads to a consistent and persistent 3D semantic scene model with the field of perception not being limited by the measurement range or occlusions of a single sensor.

To enable perception of these additional characteristics of the scene, the sensor network is extended with updated smart edge sensors with higher compute capabilities and greater flexibility\wrt the employed vision CNNs, as shown in \reffig{fig:teaser}.
This enables to run object detection and semantic image segmentation together with human pose estimation on the sensors in real time.
RGB-D cameras estimate 3D scene geometry and thermal cameras increase the person detection performance in low-light conditions.
Semantic information from detections and image segmentation is fused into the point cloud computed from the depth image and 2D human joint detections are augmented with the depth measured at the keypoint location.
Semantic point cloud and human poses are communicated to a central backend, where they are fused into an allocentric 3D metric-semantic scene model.
Only the semantic information is sent over the network; the raw images remain on the sensor boards, significantly reducing the required bandwidth, and mitigating privacy issues for the observed persons.

The semantic point clouds from multiple viewpoints are aggregated into an allocentric map of 3D scene geometry and semantic classes on the backend. The map is further updated via ray-tracing to account for moving objects.
3D human poses are estimated in real time in the scene via multi-view triangulation. The allocentric 3D human poses are projected into the local camera views and sent back to the sensors as semantic feedback~\cite{Bultmann_RSS_2021}, where they are fused with the local detections.
The 3D scene geometry enables to compute occlusion information for each joint in the respective camera view. This information is included into the semantic feedback from backend to sensors, improving the local scene model of each sensor by incorporating global context information. Unreliable, occluded joint detections can be discarded, and the local model is completed by the more reliable semantic feedback reprojected from the global, fused 3D model.

We evaluate the proposed system in experiments with challenging real-world multi-person scenes.
In summary, our contributions are:
\begin{itemize}
\item The development of a smart edge sensor platform based on the Nvidia Jetson Xavier NX development kit and an RGB-D and thermal camera, running efficient vision CNN models for object detection and semantic segmentation together with human pose estimation on-device in real time;
\item Temporal multi-view fusion of semantic point clouds from individual sensors into an allocentric semantic map of 3D scene geometry;
\item The integration of multiple instances of the proposed novel sensor nodes into a network of distributed smart edge sensors for real-time multi-view 3D human pose estimation using semantic feedback~\cite{Bultmann_RSS_2021}, complementing the feedback from backend to sensor with occlusion information for human joints in the respective camera views, computed via ray-tracing through the estimated 3D scene geometry.
\end{itemize}
We make our implementation for both sensor boards\footnote{\url{https://github.com/AIS-Bonn/JetsonTRTPerception}} and backend\footnote{\url{https://github.com/AIS-Bonn/SmartEdgeSensor3DScenePerception}} publicly available.

\section{Related Work}
\label{sec:Related_Work}
\paragraph*{Lightweight Vision CNNs for Embedded Hardware.}
Convolutional neural networks (CNNs) set the state-of-the-art for image processing and computer vision. On systems with restricted computational resources, like mobile embedded sensor platforms, however, lightweight, efficient models must be employed to achieve real-time performance. 
A popular approach is to replace classical backbone networks such as ResNets~\cite{he_deep_2016} with MobileNet~\cite{mobilenetv2_2018,mobilenetv32019} or EfficientNet~\cite{tan_efficientnet_2019} architectures, as the main computational load of CNN inference often lies in the backbone feature extractor.
These architectures decrease the number of parameters and the computational cost significantly, \eg by replacing standard convolutions with depthwise-separable convolutions.

For object detection on embedded devices, single-stage architectures such as SSD~\cite{liu_ssd_2016} or YOLO~\cite{Redmon_YOLO_2016}, which use predefined anchors instead of additional region proposal networks, were shown to be efficient. In our work, we employ the recently proposed MobileDets~\cite{xiong_mobiledets_2021}, that are optimized for embedded inference accelerators using the SSD architecture with MobileNet\,v3 backbone.

The DeepLab\,v3+ architecture~\cite{deeplabv3plus2018} for semantic segmentation uses elements of MobileNets, such as depth\-wise-separable convolutions, for efficiency on embedded hardware and shows state-of-the-art performance on large, general datasets. We employ a DeepLab\,v3+ model with MobileNet\,v3 backbone in our work.

For human pose estimation, OpenPose~\cite{cao_openpose_2018} set a new standard by detecting body parts of multiple persons in an image and associating them to individuals via Part Affinity Fields (PAFs). This bottom-up approach scales well with the number of person detections. Top-down approaches, on the other hand, first detect individuals and then estimate body keypoints for each single-person crop. These approaches achieve higher accuracy and better scale-invariance, as the person detections are interpolated to a fixed input resolution before pose inference. Xiao et al.~\cite{xiao_simple_2018} propose an efficient CNN architecture consisting of a backbone feature extractor and deconvolutional layers. We adopt this architecture and replace the ResNet backbone with MobileNet\,v3 for better efficiency on embedded hardware.

\paragraph*{Semantic Mapping.}
Semantic information about the environment is a prerequisite for many high-level robotic functions.
For this, semantic mapping systems build an allocentric model of 3D scene geometry with semantic class information.

SemanticFusion~\cite{mccormac_semanticfusion_2017} builds semantic maps from RGB-D camera input using surface elements (Surfels), where a Gaussian approximates the local point distribution. Pixel-wise class probabilities are obtained from the color image via semantic segmentation and fused into the map using a Bayesian approach assuming independence of individual measurements.

Dengler \etal\cite{dengler2021ecmr} proposed an object-centric 2D/3D map representation for real-time service robotics applications, using RGB-D data as input. A geometric segmentation of small objects in the point cloud is obtained via Euclidean clustering.
Stückler \etal\cite{stueckler_iros2012} fuse probabilistic object segmentations from multiple RGB-D camera views into a voxel-based 3D map using a Bayesian framework.

Recently, Bultmann \etal\cite{bultmann2021real} proposed  a framework for online multi-modal semantic fusion onboard a UAV combining 3D LiDAR data with detections and semantic segmentation of 2D color and thermal images. The semantic point clouds are aggregated into an allocentric voxel-based map using poses from LiDAR odometry to transform multiple viewpoints into a common coordinate frame.

\paragraph*{3D Human Pose Estimation.}
2D human joint detection, inferred by image CNNs as introduced above, provides the input for 3D human pose estimation.
3D poses are recovered from 2D keypoint detections from multiple, calibrated camera views via variants of the Pictorial Structures Model (PSM)~\cite{qiu_cross_2019,dong_fast_2019} or based on direct triangulation~\cite{chen_crossview_2020,remelli_lightweight_2020}. The PSM approaches are computationally expensive, due to a large discrete state space used in the optimization, restricting them to offline processing. %
Triangulation-based approaches are more computationally efficient and enable 3D pose estimation for multiple persons in real time.

In previous work~\cite{Bultmann_RSS_2021}, we proposed a pipeline for real-time 3D human pose estimation using multiple calibrated smart edge sensors that perform 2D pose estimation on-device. Semantic pose information is transmitted to a central backend where multiple views are fused into a 3D skeleton via triangulation and an efficient, factor graph-based skeleton model. The fused allocentric 3D joint positions, after motion prediction to compensate for the pipeline delay, are reprojected into local views and sent back to the sensors as semantic feedback, where they are fused with the detected keypoint heatmaps. This enables the sensors to incorporate global context information into their local scene view interpretation. The pipeline delay is estimated as the difference of the timestamps of the current detection and the latest received feedback message on a sensor and updated using a moving average filter.

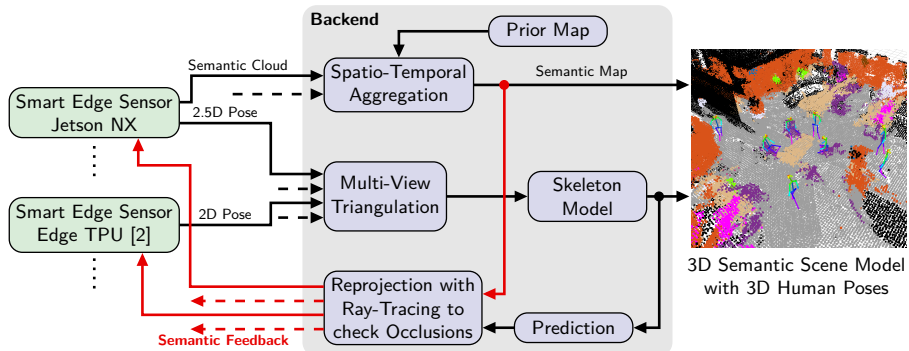
\begin{figure}[t]
  \centering
  \resizebox{1.0\linewidth}{!}{%
\begin{tikzpicture} 
[content_node/.append style={font=\sffamily,minimum size=1.5em,minimum width=6em,draw,align=center,rounded corners,scale=0.65},
label_node/.append style={font=\sffamily,scale=0.5},
group_node/.append style={font=\sffamily,dotted,align=center,rounded corners,inner sep=1em,thick},>={Stealth[inset=0pt,length=4pt,angle'=45]}]
\tikzset{junction/.append style={circle, fill=black, minimum size=3pt, draw}}

\definecolor{red}{rgb}     {0.9,0.0,0.0}
\definecolor{green}{rgb}   {0.0,0.5,0.0}
\definecolor{blue}{rgb}    {0.0,0.0,0.5}
\definecolor{grey}{rgb}    {0.5,0.5,0.5}

\draw[thick, rounded corners, grey!20!white,fill] (6.9em, 3.5em) -- (19.em, 3.5em) -- (19.em, -8em) -- (6.9em, -8em) -- cycle;
\node[font=\sffamily,scale=0.6, anchor=north west] at (6.9em, 3.5em) {\textbf{Backend}};

\node(SensorJetsonNX)[content_node,fill=green!15!white] at (0, 0) {Smart Edge Sensor\\Jetson NX};
\node(SensorEdgeTPU)[content_node,fill=green!15!white, anchor=north west, yshift = -3em] at (SensorJetsonNX.south west) {Smart Edge Sensor\\Edge TPU~\cite{Bultmann_RSS_2021}};

\node(Mapping)[content_node,fill=blue!15!white, anchor=north west, yshift = 1.5em, xshift = 7.3em] at (SensorJetsonNX.north east) {Spatio-Temporal\\Aggregation};
\node(Triang)[content_node,fill=blue!15!white, minimum height=3.5em, anchor=north west, yshift = -2.5em] at (Mapping.south west) {Multi-View\\Triangulation};
\node(Reproj)[content_node,fill=blue!15!white, anchor=north west, yshift = -2.em] at (Triang.south west) {Reprojection with\\Ray-Tracing to\\check Occlusions};

\node(SkelModel)[content_node,fill=blue!15!white, anchor=west, xshift = 4.1em] at (Triang.east) {Skeleton\\Model};
\node(Pred)[content_node,fill=blue!15!white, anchor=north west, xshift = -.7em, yshift = -4.6em] at (SkelModel.south west) {Prediction};
\node(PriorMap)[content_node,fill=blue!15!white, anchor=south west, xshift=-2em, yshift = 6.3em] at (SkelModel.north west) {Prior Map};

\node(out)[anchor=south west, xshift=1.1em, yshift=-1.2em] at (SkelModel.south east){\includegraphics[width=2.3cm,trim=80 0 70 0, clip]{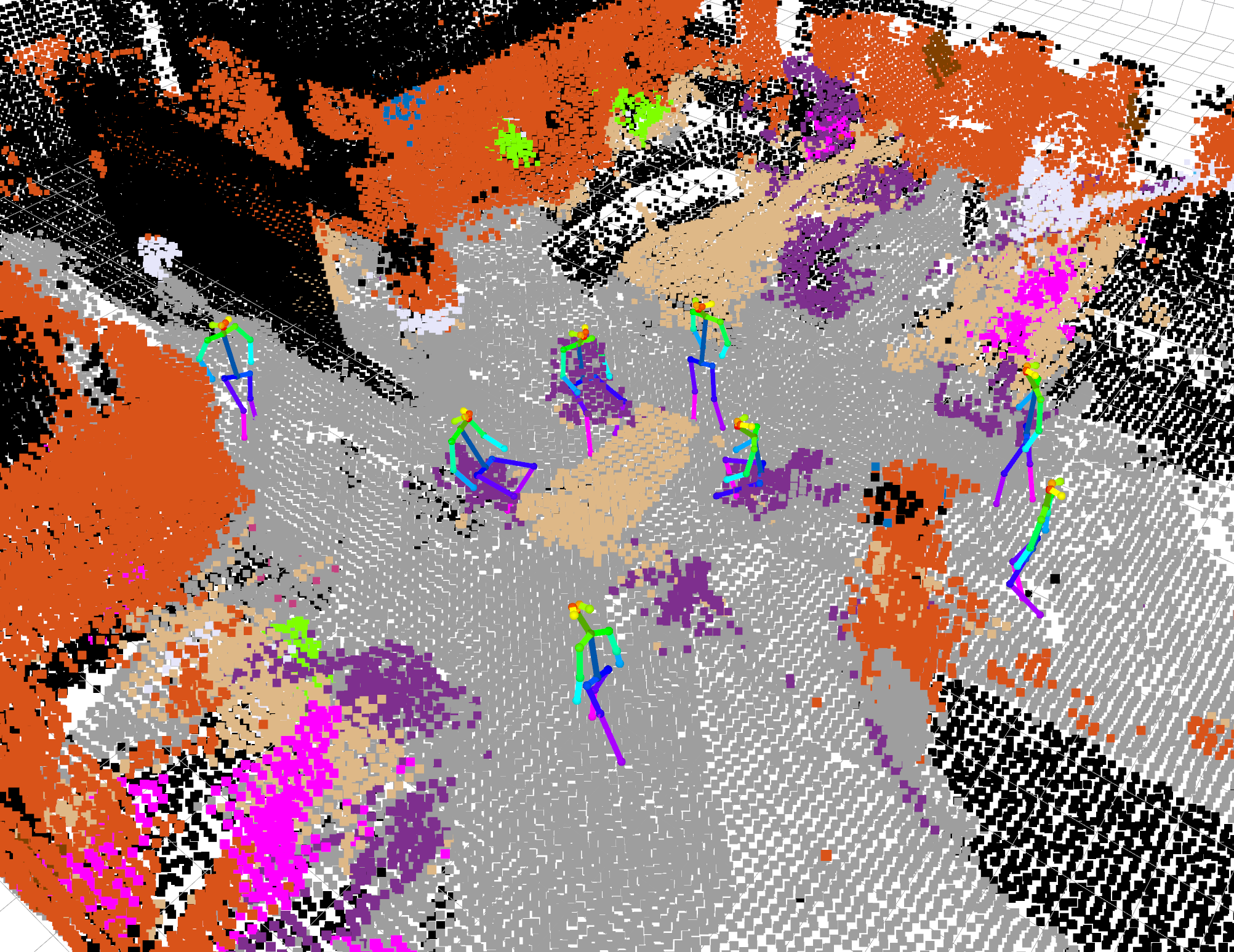}};
\node(out_text)[font=\sffamily,align=center,scale=0.65,anchor=north west, yshift=.4em] at (out.south west) {3D Semantic Scene Model\\with 3D Human Poses};

\draw[dotted, thick] (SensorJetsonNX) ++(0, -1.1em) -- ++(0, -1.2em);
\draw[dotted, thick] (SensorEdgeTPU) ++(0, -1.1em) -- ++(0, -1.2em);

\draw[->, thick] (SensorJetsonNX.7) -| (3.2em, 0 |- Mapping.173) -- ++(0.6em, 0) -- node[label_node,near start,above] {Semantic Cloud} (Mapping.173);
\draw[->, dashed, thick] (Mapping.187) + (-3.em,0) -- (Mapping.187);
\draw[->, thick] (PriorMap.180) -| (Mapping.90);
\draw[->, thick] (Mapping.0) -- node[label_node,midway,above] {Semantic Map} ++(7.1em, 0);

\draw[->, thick] (SensorJetsonNX.353) -- node[label_node,midway,above] {2.5D Pose} ++(3em, 0) |- (Triang.160);
\draw[->, dashed, thick] (Triang.173) + (-1.5em,0) -- (Triang.173);
\draw[->, thick] (SensorEdgeTPU.0) -- node[label_node,midway,above] {2D Pose} ++(3em, 0) |- (Triang.186);
\draw[->, dashed, thick] (Triang.199) + (-1.5em,0) -- (Triang.199);
\draw[->, thick] (Triang.0) -- (SkelModel.180);
\draw[->, thick] (SkelModel.0) -- ++(1.4em, 0);

\draw[->, thick] (SkelModel.0) + (.4em, 0) |- (Pred.0);
\draw[->, thick] (Pred.180) -- (Reproj.0 |- Pred.180);

\draw[->, thick, red] (Mapping.0) + (1em, 0) |- (Reproj.10);
\draw[->, thick, red] (Reproj.165) -- ++(-4.4em, 0) |- ++(0, 4em) -| (SensorJetsonNX.330);
\draw[->, dashed, thick, red] (Reproj.175) -- ++(-4.4em, 0);
\draw[->, thick, red] (Reproj.185) -| (SensorEdgeTPU.330);
\draw[->, dashed, thick, red] (Reproj.195) -- node[label_node,near end,below] {\textbf{Semantic Feedback}} ++(-4.4em, 0);

\node(junct_0)[junction, red, anchor=center, scale=0.3] at (Mapping.0 -| 13.5em, 0){};
\node(junct_1)[junction, anchor=center, scale=0.3] at (SkelModel.0 -| 18.6em, 0){};

\end{tikzpicture}
}
  \caption{Overview of the multi-sensor pipeline for 3D semantic mapping and human pose estimation: The Jetson NX smart edge sensors extend a sensor network from prior work~\cite{Bultmann_RSS_2021} of nodes with lower compute capabilities. Semantic point clouds from multiple sensor views are aggregated into an allocentric 3D semantic map and 3D human poses are estimated in real time. The map is used to check reprojected joints for occlusion in the resp. sensor view via ray-tracing and this information is added to the semantic feedback sent to the smart edge sensors.}
  \label{fig:sensor_network}
  \vspace{-1em}
\end{figure}
We build upon this work and extend the sensor network with new smart edge sensor nodes with significantly increased computational power and RGB-D cameras that enable the perception of 3D geometry.
In addition to human pose estimation, object detection and semantic image segmentation are computed on the sensor boards and fused via 3D projection into a semantic point cloud.
Semantic point clouds from multiple sensor views are fused into a sparse voxel hash-map with per-voxel full semantic class probabilities on the backend.
The semantic map is used to obtain occlusion information for person keypoints in the local camera views, which is added to the semantic feedback to increase the robustness of the pose estimation pipeline.

\section{Method}
\label{sec:method}
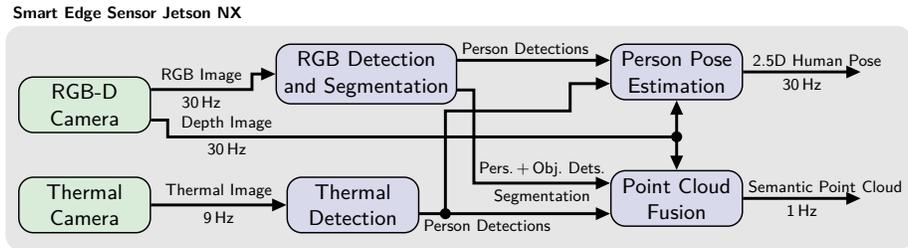
\begin{figure}[t]
  \centering
  \resizebox{1.0\linewidth}{!}{%
\begin{tikzpicture} 
[content_node/.append style={font=\sffamily,minimum size=1.5em,minimum width=6em,draw,align=center,rounded corners,scale=0.65},
label_node/.append style={font=\sffamily,scale=0.5},
group_node/.append style={font=\sffamily,dotted,align=center,rounded corners,inner sep=1em,thick},>={Stealth[inset=0pt,length=4pt,angle'=45]}]
\tikzset{junction/.append style={circle, fill=black, minimum size=3pt, draw}}

\definecolor{red}{rgb}     {0.5,0.0,0.0}
\definecolor{green}{rgb}   {0.0,0.5,0.0}
\definecolor{blue}{rgb}    {0.0,0.0,0.5}
\definecolor{grey}{rgb}    {0.5,0.5,0.5}

\draw[thick, rounded corners, grey!20!white,fill] (-2.3em,2.3em) -- (24.5em,2.3em) -- (24.5em,-4.3em) -- (-2.3em,-4.3em) -- cycle;

\node(Camera)[content_node,fill=green!15!white] at (0, 0) {RGB-D\\Camera};
\node(Thermal_Camera)[content_node,fill=green!15!white, anchor=north west, yshift = -2em] at (Camera.south west) {Thermal\\Camera};

\node(RGB_Segm_Detection)[content_node,fill=blue!15!white, anchor=north west, yshift = 1.5em, xshift = 5.7em] at (Camera.north east) {RGB Detection\\and Segmentation};
\node(Thermal_Detection)[content_node,fill=blue!15!white, anchor=north west, xshift = 6.2em] at (Thermal_Camera.north east) {Thermal\\Detection};

\node(Pose_Estimation)[content_node,fill=blue!15!white, anchor=north west, xshift = 7em] at (RGB_Segm_Detection.north east) {Person Pose\\Estimation};
\node(Pcd_Fusion)[content_node,fill=blue!15!white, anchor=north west, yshift = -3.2em] at (Pose_Estimation.south west) {Point Cloud\\Fusion};

\draw[->, thick] (Camera.13) -- node[label_node,midway,below] {\SI{30}{\hertz}} node[label_node,midway,above] {RGB Image} ++(3em, 0) |- (RGB_Segm_Detection);
\draw[->, thick] (Camera.347) -| ++(0.5em, -.5em) -- node[label_node,midway,below] {\SI{30}{\hertz}} node[label_node,midway,above] {Depth Image} ++(3.5em, 0) -| (Pcd_Fusion.90);
\draw[->, thick] (Thermal_Camera) -- node[label_node,midway,below] {\SI{9}{\hertz}} node[label_node,midway,above] {Thermal Image} (Thermal_Detection);
\draw[->, thick] (Pcd_Fusion.90 |- 0.,-1em) -- (Pose_Estimation.270);

\draw[->, thick] (RGB_Segm_Detection.10 |- Pose_Estimation.170) -- node[label_node,midway,above,xshift=-.5em] {Person Detections} (Pose_Estimation.170);
\draw[->, thick] (RGB_Segm_Detection.350) -| ++(0.5em, -2em) |- node[label_node,near end,above] {Pers.\,+\,Obj. Dets.} node[label_node,near end,below] {Segmentation} (Pcd_Fusion.167);
\draw[->, thick] (Thermal_Detection.347 |- Pcd_Fusion.193) -- node[label_node,near start,below,xshift=1.2em] {Person Detections} (Pcd_Fusion.193);
\draw[->, thick] (10.7em, 0 |- Pcd_Fusion.193) -- ++(0, 3.05em) -- ++(3.7em, 0) |- (Pose_Estimation.190);

\draw[->, thick] (Pose_Estimation.0) -- node[label_node,midway,above,xshift=.9em] {2.5D Human Pose} node[label_node,midway,below] {\SI{30}{\hertz}} ++(3.5em, 0);
\draw[->, thick] (Pcd_Fusion.0) -- node[label_node,midway,above,xshift=1.4em] {Semantic Point Cloud} node[label_node,midway,below] {\SI{1}{\hertz}} ++(3.5em, 0);

\node(junct_0)[junction, anchor=center, scale=0.3] at (Pcd_Fusion.90 |- 0.,-.95em){};
\node(junct_1)[junction, anchor=center, scale=0.3] at (10.7em, 0 |- Pcd_Fusion.193){};

\node(PC_Group_Label)[label_node,anchor=south west] at (-2.3em, 2.3em) {\textbf{Smart Edge Sensor Jetson NX}};
\end{tikzpicture}
}
  \caption{Smart edge sensor semantic perception system overview. Human poses are estimated in real time, while the semantic point cloud of the static or slowly moving scene geometry is output at a lower frequency to save compute resources.}
  \label{fig:system}
\end{figure}
Figure~\ref{fig:sensor_network} illustrates the proposed multi-sensor pipeline for 3D semantic perception and human pose estimation combining two types of smart edge sensors.
The proposed Jetson~NX sensors are integrated into a sensor network from prior work~\cite{Bultmann_RSS_2021}, consisting of nodes based on the Google Edge~TPU with lower compute capabilities and RGB image-only 2D human pose estimation, without local depth estimation.
We consider a calibrated camera network, with known projection matrices from sensor to world coordinates, where the sensors are software-synchronized via NTP.
Semantic mapping is only performed with the here proposed Jetson~NX sensor nodes, while data from both sensor types is combined for 3D human pose estimation.

An overview of the proposed approach for semantic perception onboard each Jetson~NX smart edge sensor is given in \reffig{fig:system}.
We detail individual components of the data processing on each sensor board, as well as the fusion of multiple sensor views for 3D mapping and 3D human pose estimation in the following.

\subsection{Smart Edge Sensor Hardware}
\label{sec:sensor_system}
We developed smart edge sensors based on the Nvidia Jetson Xavier NX developer kit\footnote{\url{https://developer.nvidia.com/embedded/jetson-xavier-nx-devkit}} (\cf\reffig{fig:teaser}~(a)).
They are equipped with a 6-core ARM processor, 384 CUDA cores, and \SI{8}{\giga\byte} of RAM. The Jetson NX embedded system achieves a CNN inference performance of 21 trillion operations per second (TOPS), a significant increase compared to the 4 TOPS of the sensor platform employed in our previous work~\cite{Bultmann_RSS_2021}.
For visual perception, we connect an Intel RealSense D455 RGB-D camera and a FLIR Lepton 3.5 thermal camera to the Jetson NX board.

\subsection{Single-View Embedded Semantic Perception}
\label{sec:semantic_perception}
\paragraph*{Person and Object Detection.}
We employ the recent MobileDet architecture~\cite{xiong_mobiledets_2021} for person and object detection. The RGB detector is trained on the COCO dataset~\cite{lin_coco_2014} using \textit{person} and 12 indoor object classes (e.g., \textit{chair}, \textit{table}, \textit{computer\,/\,tv}), with an input resolution of 848$\,\times\,$480~px.
The same network architecture is used for the thermal detector, taking one-channel 8-bit gray-scale thermal images at the camera resolution of 160$\,\times\,$120~px as input. The thermal detector is trained on the ChaLearn IPHD dataset~\cite{flir_iphd_2020}, with annotations for the \textit{person} class only.

\paragraph*{Person Keypoint Estimation.}
We adopt a top-down approach for person pose estimation on the smart edge sensors, where crops of single persons are analyzed by the keypoint estimation CNN.
The CNN architecture of Xiao et al.~\cite{xiao_simple_2018} is the basis of our person pose estimation, but we exchange the ResNet backbone with the significantly more lightweight MobileNet\,v3 feature extractor~\cite{mobilenetv32019}. We train the pose estimation network on the COCO dataset~\cite{lin_coco_2014} using person keypoint annotations.

Person detections from RGB and thermal images are forwarded to the keypoint estimation CNN. The RGB-D depth thereby is used to project detections from the thermal camera to the color image. Redundant detections of the same person in both modalities are filtered via non-maximum suppression (NMS).
Each person crop is then resized to the fixed 192$\,\times\,$256 input resolution of the keypoint estimation CNN and inference is run for all crops together in batched mode. Batch processing gives a significant improvement in the scaling of inference time with the number of persons compared to previous work~\cite{Bultmann_RSS_2021}, where the embedded hardware only supported processing a single crop at a time (cf. \refsec{sec:runtime}).

The pose estimation model outputs multi-channel images, called \textit{heatmaps}, encoding the confidence of a joint being present at the pixel location. As single-person crops are processed, 2D joint locations are determined as global maxima of the respective heatmap channel.
The RGB-D range image is used to augment the 2D keypoints to a 2.5D pose representation. For each joint, the median depth of a 5$\,\times\,$5~px region around the joint location is obtained from the depth image.
The local depth estimate enables the projection of keypoints into three-dimensional space but often suffers from noise and occlusions, as is further analyzed in \refsec{sec:pose_est_3d}.
The 2.5D pose estimate for each detected person is sent to a central backend, where multiple sensor views are fused into a coherent 3D pose representation.
The pose estimation pipeline runs with the highest real-time priority on the sensor boards, to enable tracking of dynamic human motions.
To save computational resources, the person detector is run only once per second and the crops are updated based on the keypoint estimations between detector runs.

\paragraph*{Semantic Segmentation.}
We adopt the DeepLab\,v3+~\cite{deeplabv3plus2018} architecture with MobileNet\,v3~\cite{mobilenetv32019} backbone for semantic segmentation. We train the model on the indoor scenes of the ADE20K dataset~\cite{zhou2019semantic} and reduce the labels to 16 classes most relevant for the intended indoor application scenarios (\cf\reffig{fig:teaser}).
The input image size is set to 849$\,\times\,$481~px during inference, fitting to the 16:9 aspect ratio of our camera.
The semantic segmentation is run only once per second, similar to the person and object detector, to save computational resources.

\subsection{Multi-Modal Semantic Point Cloud Fusion}
We obtain a geometric point cloud by projecting the RGB-D range image into 3D. The point cloud is uniformly subsampled using a voxel-grid filter with \SI{5}{\centi\meter} resolution to reduce the amount of data, economizing computational resources, and later network bandwidth for transmission to the backend.
Sparse outlier measurements are further removed by a statistical outlier filter, as implemented in the PCL library~\cite{pcl_icra_2011}. A point is deleted when the distance to its neighbors is outside an interval defined by the mean and standard deviation of the entire point cloud.

Semantic information from RGB and thermal detections, as well as RGB semantic segmentation, is fused into the point cloud using a projection-based approach as proposed by Bultmann et al.~\cite{bultmann2021real}. For this, the points are projected into the segmentation mask inferred from the RGB image.
The semantic class scores $\vec{c}_{\text{segm}} \in\mathbb{R}^C$ are obtained from semantic segmentation via bilinear interpolation at the projected point location.
A normalized probability distribution over the employed $C=16$ classes is then approximated by applying the soft-max operation:
\begin{align}
p_i &= \sigma\left( c_i \right) = \frac{\exp{c_i}}{\sum^C_{j=1}\exp{c_j}}\,,
\end{align}
obtaining $\vec{p}_\text{segm}\in\mathbb{R}^{C}$, with $p_i \in [0,1]$ and $\sum_i p_i = 1$.
If a projected point falls inside a detection bounding box in either thermal or color images, we further fuse the detector result with the semantic segmentation.
We reconstruct the detection probability distribution $\vec{p}_\text{det}$ from the score for the detected class following the maximum entropy principle:
The probability of the detected class $p_\text{det}$ is given by the detector score and the remaining probability mass $1-p_\text{det}$ is equally distributed over the remaining $C-1$ classes.
Both estimates are fused following the Bayesian update rule~\cite{mccormac_semanticfusion_2017}, assuming independence of segmentation and detection:
\begin{align}
\vec{p}_\text{fused} &= \frac{\vec{p}_\text{segm} \circ \vec{p}_\text{det}}{\sum_{i=1}^C p_{i,\text{segm}}\,p_{i,\text{det}}}\,,
\end{align}
with $\circ$ being the coefficient-wise product. For better numerical stability, we use the implementation of the Bayesian fusion in logarithmic form from~\cite{bultmann2021real}.

As the detection bounding boxes are axis-aligned, border-effects have to be considered for non-rectangular or non-axis-aligned objects before detection fusion.
Inclusion of all points projected into the bounding box in the fusion would falsely label points on the ground and in the background as the detected class. To alleviate this issue, the ground plane is removed and the remaining points are clustered in 3D Euclidean space by a distance threshold~\cite{bultmann2021real}. Only the clustered points are included into detection fusion.

The output semantic point cloud includes the class probability vector and the argmax class color per point (\cf\reffig{fig:semantic_map}~(a)) and is sent to the central backend over the network. It is computed at a reduced update frequency of \SI{1}{\hertz} on the sensors, as it is targeted to observe the static or slowly moving scene geometry. Thus, computational resources are kept free for the real-time estimation of dynamic human motions in the pose estimation pipeline.
 
\subsection{3D Semantic Mapping}
\label{sec:sematic_map}
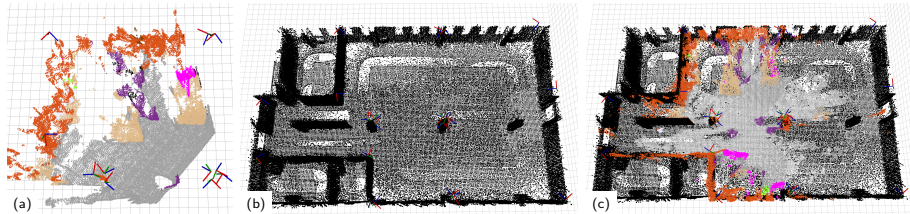
\begin{figure}[t]
	\centering
		\begin{tikzpicture}
			\node[inner sep=0,anchor=north west] (image1) at (0, 0) {\includegraphics[height=2.85cm,trim=200 240 380 20, clip]{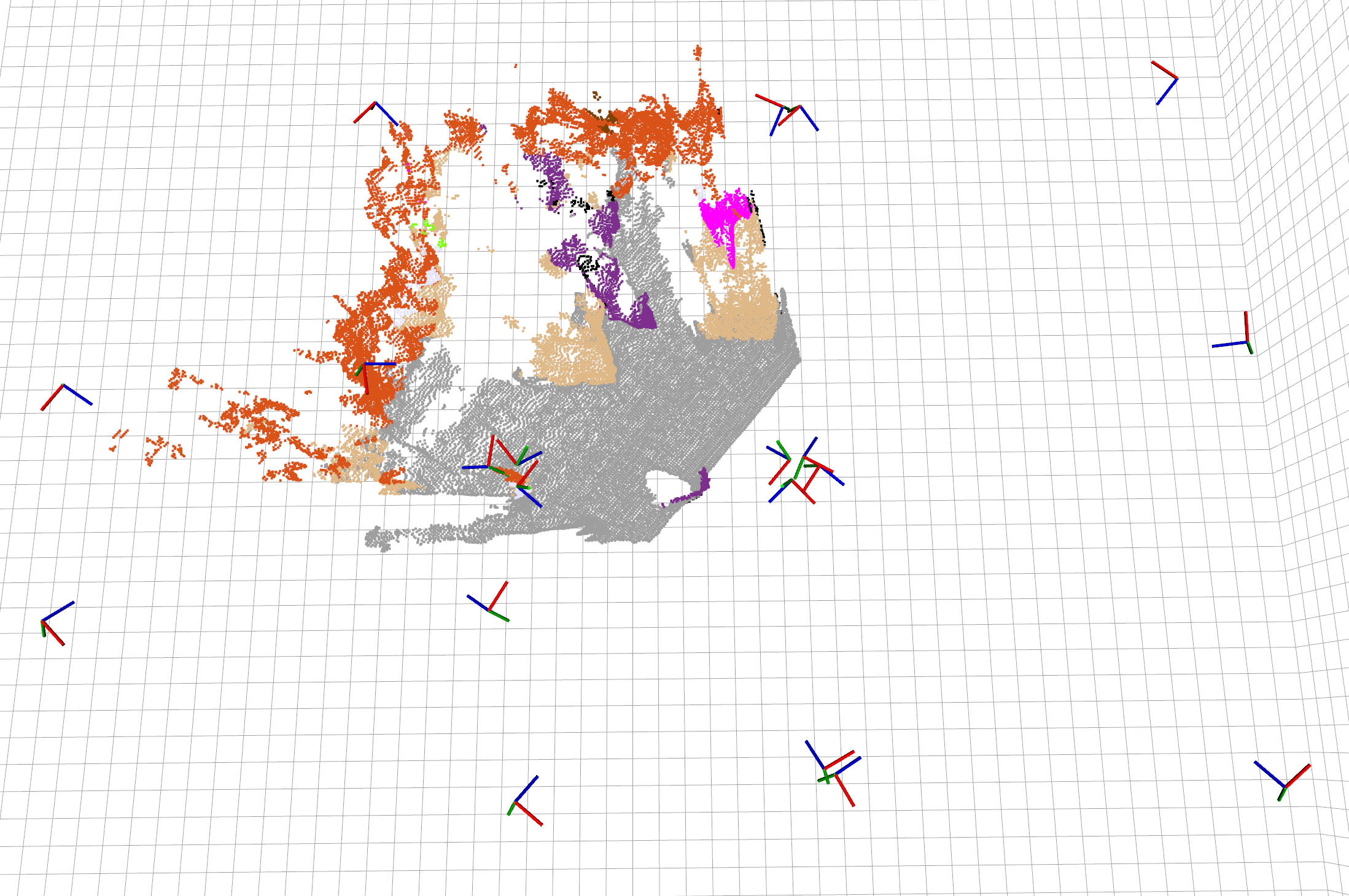}};
			\node[inner sep=0,anchor=north west,xshift=0.1cm] (image2) at (image1.north east) {\includegraphics[height=2.85cm,trim=0 0 0 0, clip]{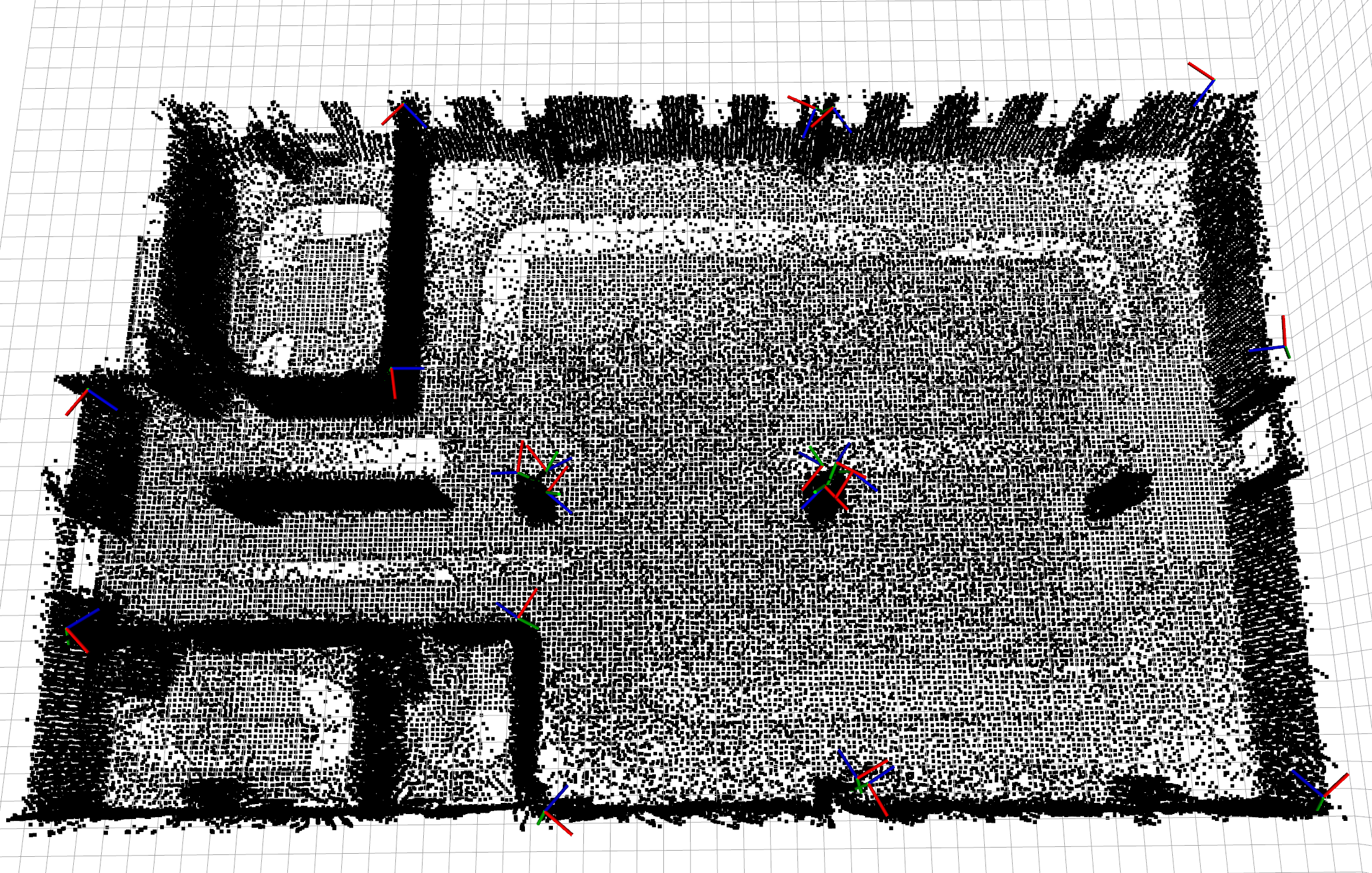}};
			\node[inner sep=0,anchor=north west,xshift=0.1cm] (image3) at (image2.north east) {\includegraphics[height=2.85cm,trim=0 0 0 0, clip]{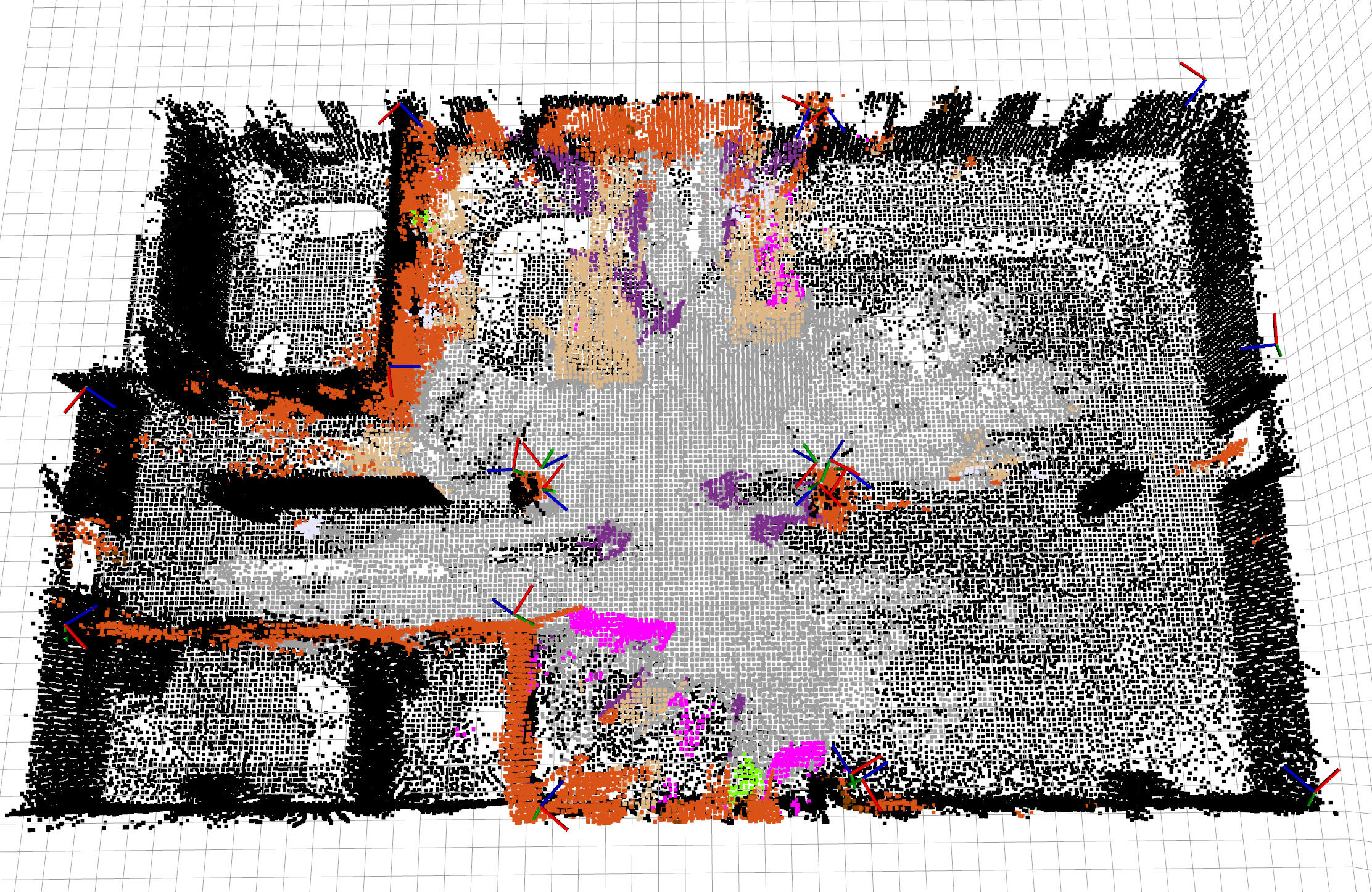}};
			
			\node[label,scale=.75, anchor=south west, rectangle, fill=white, align=center, font=\scriptsize\sffamily] (n_0) at (image1.south west) {(a)};
			\node[label,scale=.75, anchor=south west, rectangle, fill=white, align=center, font=\scriptsize\sffamily] (n_1) at (image2.south west) {(b)};
			\node[label,scale=.75, anchor=south west, rectangle, fill=white, align=center, font=\scriptsize\sffamily] (n_1) at (image3.south west) {(c)};
		\end{tikzpicture}
	\caption{3D Semantic Mapping: (a) semantic point cloud of a single sensor, (b) prior map, (c) fused semantic map. The smart edge sensors send semantic point clouds of their resp. perspectives to the backend. Here, the fused map is initialized with a prior map and updated with the observations including semantic classes.}
	\label{fig:semantic_map}
\end{figure}
Semantic point clouds from multiple calibrated camera perspectives are fused into an allocentric semantic map on the central backend, as illustrated in \reffig{fig:semantic_map}. For this, the 3D space is uniformly subdivided into cubic volume elements (voxels). We employ sparse voxel hashing~\cite{quenzel2021mars} as a memory-efficient data structure.

For indoor environments, prior information on building structure is often easily available, e.g., via floor plans or 3D models. To incorporate this prior information, we initialize the scene model with a prior map of the empty building (\reffig{fig:semantic_map}~(b)). Here, the prior map was obtained from aggregated laser scans of the empty rooms, but it could also be replaced, e.g., by a floor plan with a fixed wall height or an architectural CAD model of the building.

To include semantic information and current observations into the map, we transform the semantic point clouds from individual sensors (\reffig{fig:semantic_map}~(a)) into global coordinates using the known camera calibration and bin the points into voxels of \SI{10}{\centi\meter} side length.
The semantic probabilities of all points falling into a voxel are fused probabilistically, using Bayes' rule~\cite{mccormac_semanticfusion_2017}, assuming independence between observations $P\left( l_i \lvert X_k\right)$ for the semantic point cloud $X_k$ with label $l_i$ for class $i$:
\begin{align}
P( l_i \lvert X_{1:k} ) &=  \frac{P\left(l_i \lvert X_{1:k-1}\right)  P \left( l_i \lvert X_k \right)}{\sum_i P\left( l_i \vert X_{1:k-1} \right) P\left( l_i \vert X_k\right)}\,.\label{eq:bayes}
\end{align}
We again use the implementation of Bayesian fusion in logarithmic form from~\cite{bultmann2021real} for better numerical stability.
Points labeled as \textit{person} are not included in the semantic map, as dynamic human segments are tracked at a higher update rate via the 3D skeleton representation (\cf\refsec{sec:pose_est_3d}).

The fused semantic map (\reffig{fig:semantic_map}~(c)) contains 3D geometry and semantic classes of the areas observed by the smart edge sensors and is completed by the prior information for currently unobserved areas.

\begin{figure}[t]
	\centering
		\begin{tikzpicture}
			\node[inner sep=0,anchor=north west] (image1) at (0, 0) {\includegraphics[height=3.2cm,trim=0 0 0 0, clip]{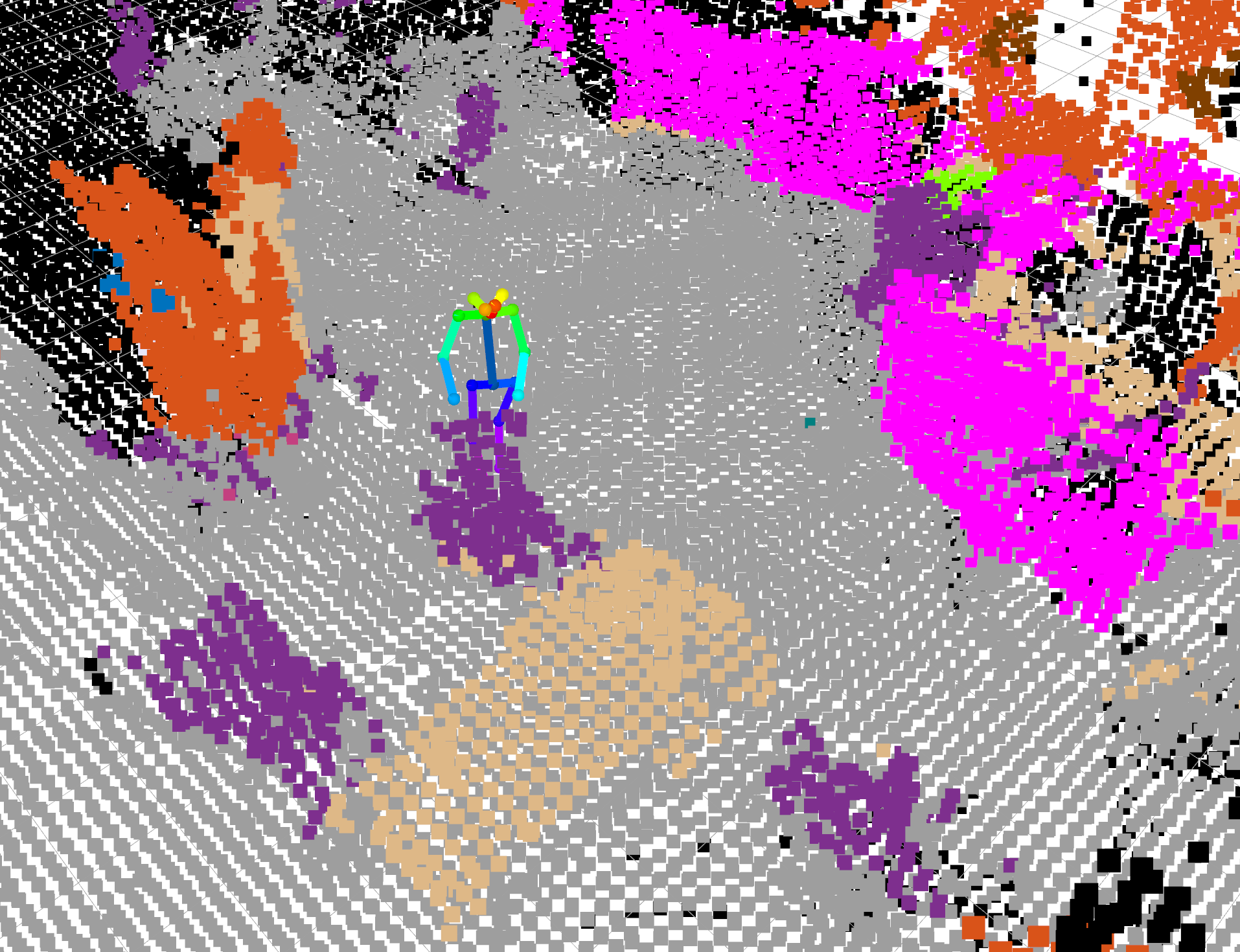}};
			\node[inner sep=0,anchor=north west,xshift=0.2cm] (image2) at (image1.north east) {\includegraphics[height=3.2cm,trim=0 0 70 0, clip]{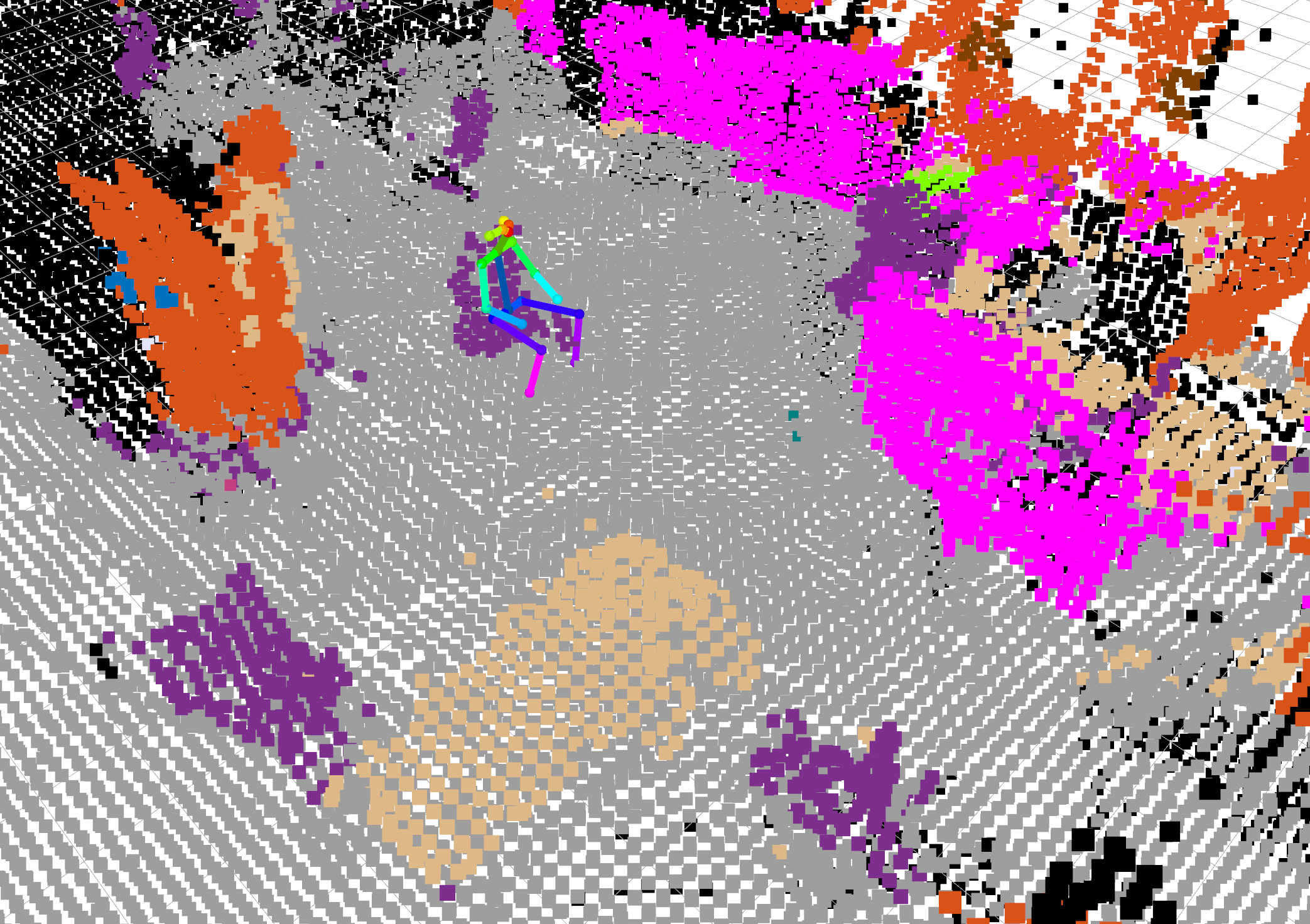}};
			
			\begin{scope}[shift=(image1.north west),x={(image1.north east)},y={(image1.south west)}]
				\draw[red!80, ultra thick] (0.4,0.46) ellipse (0.12 and 0.22);
			\end{scope}
			\begin{scope}[shift=(image2.north west),x={(image2.north east)},y={(image2.south west)}]
				\draw[red!80, ultra thick] (0.42,0.33) ellipse (0.11 and 0.14);
			\end{scope}
			
			\node[label,scale=.75, anchor=south west, rectangle, fill=white, align=center, font=\scriptsize\sffamily] (n_0) at (image1.south west) {(a)};
			\node[label,scale=.75, anchor=south west, rectangle, fill=white, align=center, font=\scriptsize\sffamily] (n_1) at (image2.south west) {(b)};
		\end{tikzpicture}
	\caption{Map update: 3D Semantic map and 3D person skeleton (a) before and (b) after moving a chair (highlighted with red circle). The semantic map is updated via ray-tracing to account for moving objects.}
	\label{fig:map_update}
\end{figure}
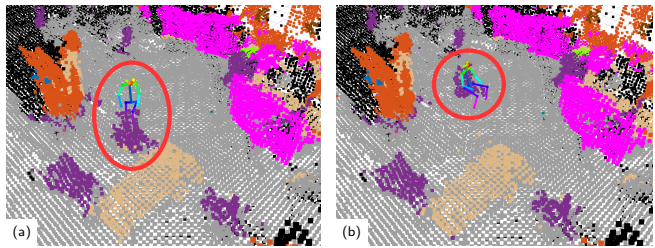
To account for moving objects, we adopt a simple ray-tracing approach to update occupancy information of the voxels~\cite{schleich_icuas_2021}, as illustrated in \reffig{fig:map_update}. Starting from the sensor pose towards the measured voxels, we ray-trace using a 3D implementation of Bresenham's algorithm~\cite{bresenham3D1987}. All voxels between the start and endpoint of the ray are updated as being free space, while the measured voxels are updated as being occupied. The semantic class probability is reset when a voxel state transitions from occupied to free.

\subsection{3D Human Pose Estimation with Occlusion Feedback}
\label{sec:pose_est_3d}
The 3D joint positions of the detected persons are recovered from a set of 2D keypoint detections from multiple viewpoints via triangulation, and the result is refined using a factor graph skeleton model, as proposed in~\cite{Bultmann_RSS_2021}.
Furthermore, a semantic feedback channel from backend to sensors is implemented in our framework that enables the local semantic models of each sensor to incorporate globally-fused 3D pose information.

In this work, we add occlusion information for each joint to the semantic feedback, using the estimated 3D semantic map (\cf\reffig{fig:sensor_network}).
We employ ray-tracing to check each joint for occlusion in the respective local sensor view. For this, we traverse the ray from the respective camera pose to the 3D joint through the 3D map using Bresenham's 3D line-search~\cite{bresenham3D1987}. When the ray hits a minimum number of $k=2$ occupied voxels, the joint is marked as occluded in the respective local view.
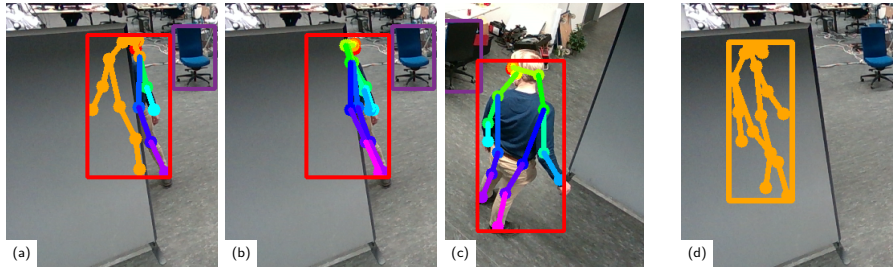
\begin{figure}[ht]
	\centering
		\begin{tikzpicture}
			\node[inner sep=0,anchor=north west] (image1) at (0, 0) {\includegraphics[height=3.5cm,trim=45 45 225 25, clip]{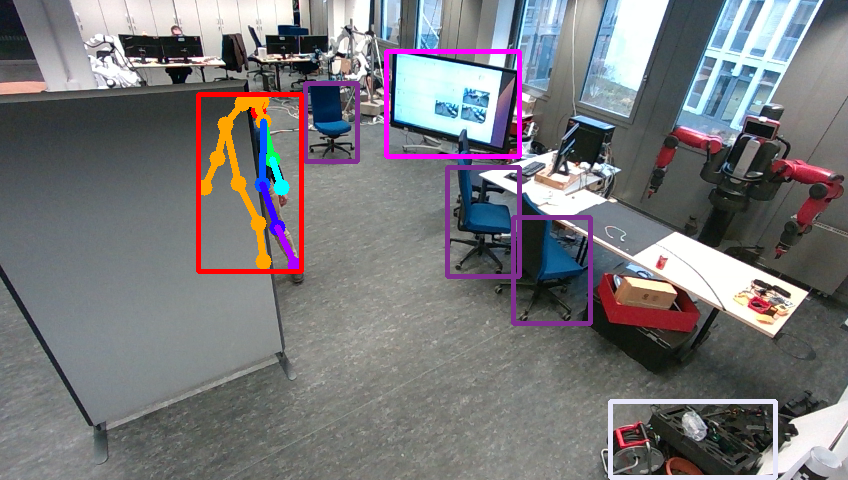}};
			\node[inner sep=0,anchor=north west,xshift=0.1cm] (image2) at (image1.north east) {\includegraphics[height=3.5cm,trim= 45 45 225 25, clip]{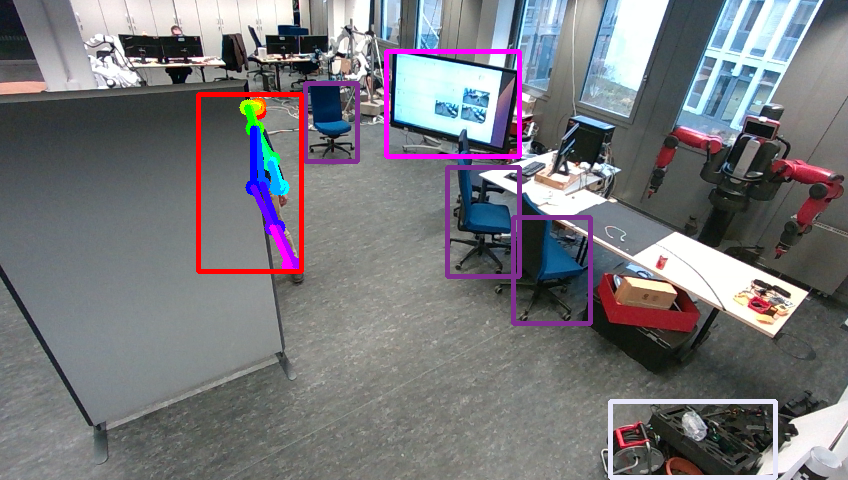}};
			\node[inner sep=0,anchor=north west,xshift=0.1cm] (image3) at (image2.north east) {\includegraphics[height=3.5cm,trim= 190 0 80 60, clip]{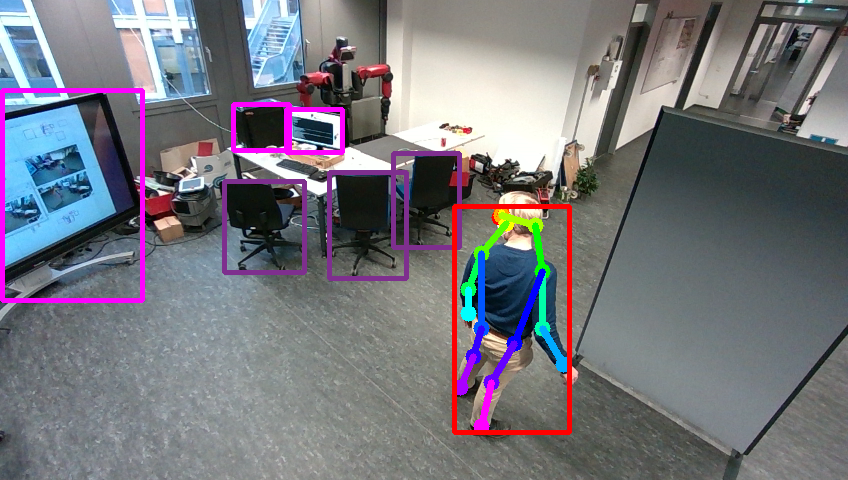}};
			\node[inner sep=0,anchor=north west,xshift=0.5cm] (image4) at (image3.north east) {\includegraphics[height=3.5cm,trim= 40 45 230 25, clip]{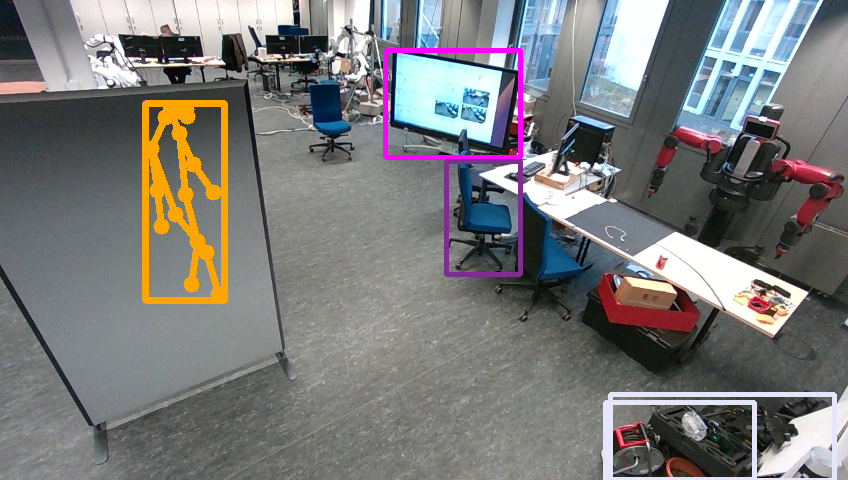}};
			
			\node[label,scale=.75, anchor=south west, rectangle, fill=white, align=center, font=\scriptsize\sffamily] (n_0) at (image1.south west) {(a)};
			\node[label,scale=.75, anchor=south west, rectangle, fill=white, align=center, font=\scriptsize\sffamily] (n_1) at (image2.south west) {(b)};
			\node[label,scale=.75, anchor=south west, rectangle, fill=white, align=center, font=\scriptsize\sffamily] (n_2) at (image3.south west) {(c)};
			\node[label,scale=.75, anchor=south west, rectangle, fill=white, align=center, font=\scriptsize\sffamily] (n_3) at (image4.south west) {(d)};
		\end{tikzpicture}
	\caption{Occlusion information in semantic feedback: Local 2D pose estimation (a) with and (b) without occlusion information via semantic feedback, (c) reference view without occlusion, (d) fully occluded person. Person detections in red and skeleton keypoints colored by joint index. Occluded joints are marked in orange.
	Heavy occlusion causes the pose estimation to collapse to the visible side only. With occlusion information, unreliable, occluded joint detections can be discarded, and the local model is completed by the more reliable semantic feedback.}
	\label{fig:occlusion}
\end{figure}
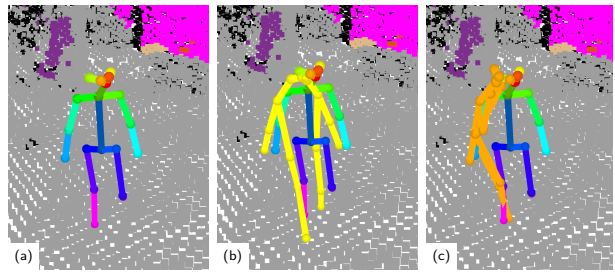
\begin{figure}[ht]
	\centering
		\begin{tikzpicture}
			\node[inner sep=0,anchor=north west] (image1) at (0, 0) {\includegraphics[height=3.5cm,trim= 20 0 0 0, clip]{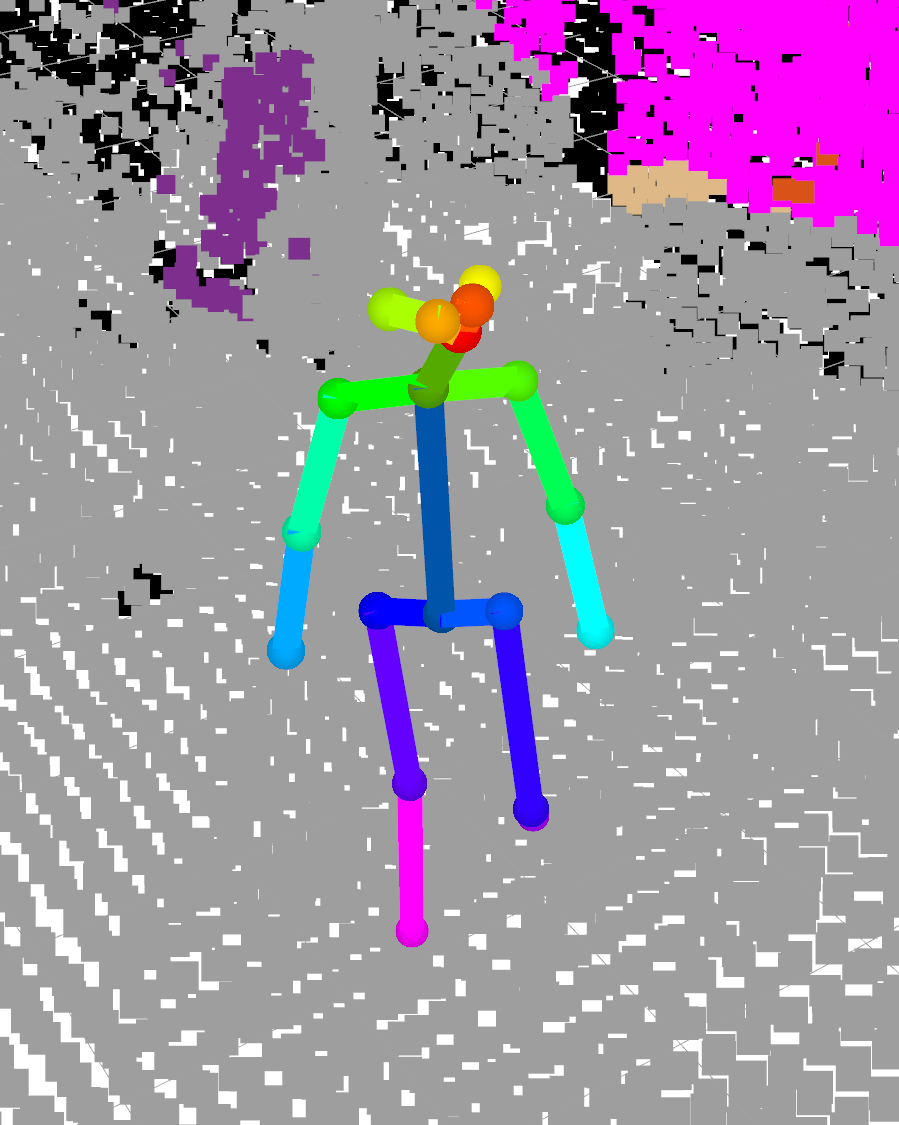}};
			\node[inner sep=0,anchor=north west,xshift=0.1cm] (image2) at (image1.north east) {\includegraphics[height=3.5cm,trim= 0 0 10 0, clip]{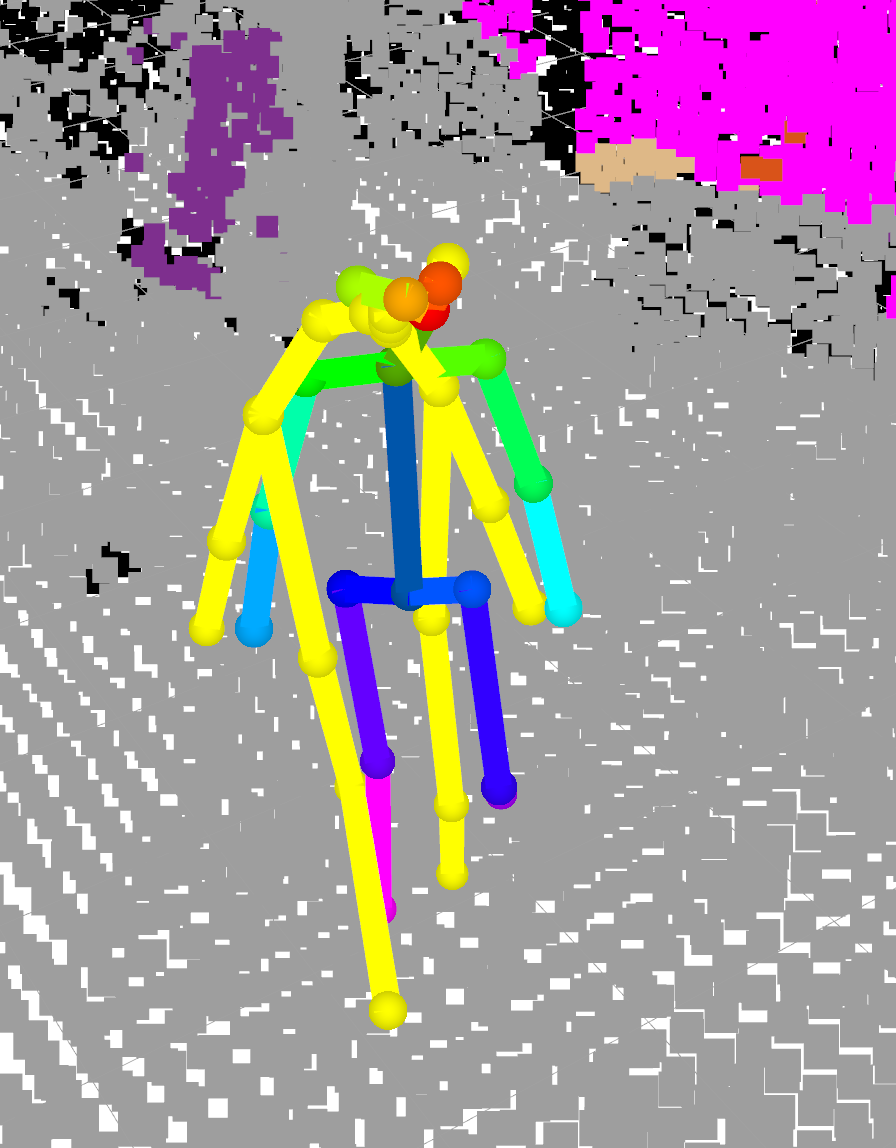}};
			\node[inner sep=0,anchor=north west,xshift=0.1cm] (image3) at (image2.north east) {\includegraphics[height=3.5cm,trim= 0 0 0 0, clip]{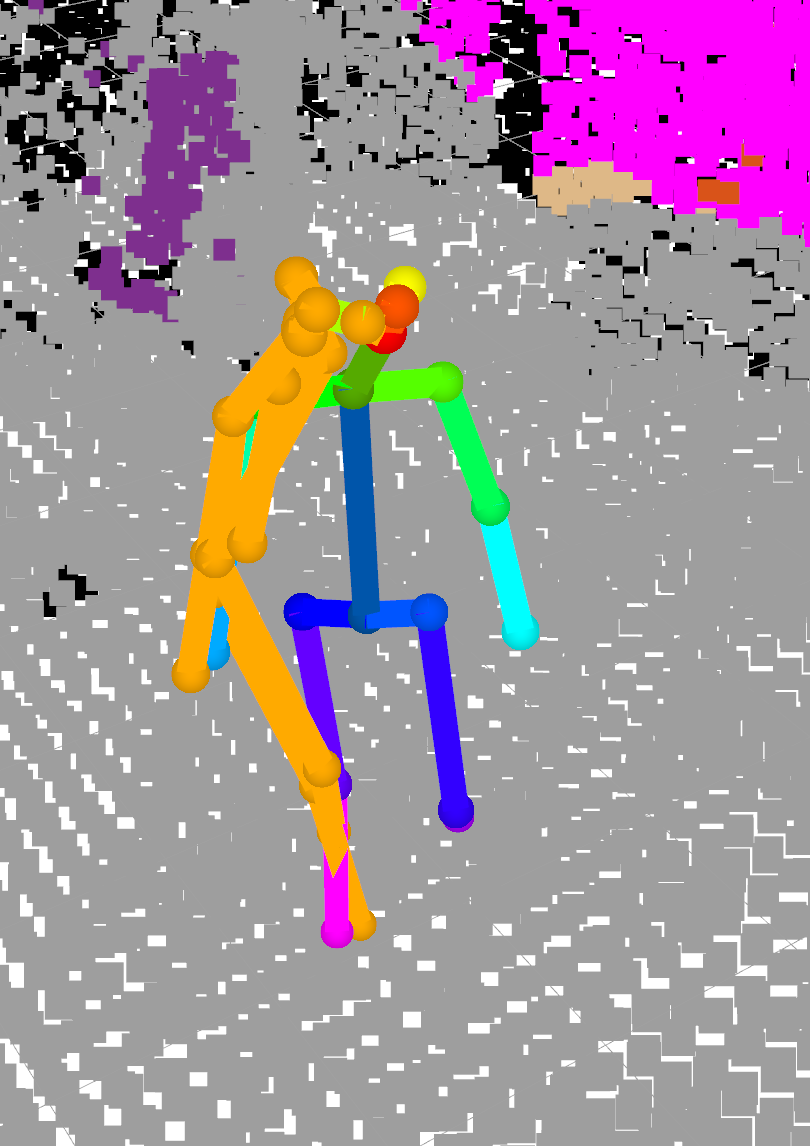}};

			\node[label,scale=.75, anchor=south west, rectangle, fill=white, align=center, font=\scriptsize\sffamily] (n_0) at (image1.south west) {(a)};
			\node[label,scale=.75, anchor=south west, rectangle, fill=white, align=center, font=\scriptsize\sffamily] (n_1) at (image2.south west) {(b)};
			\node[label,scale=.75, anchor=south west, rectangle, fill=white, align=center, font=\scriptsize\sffamily] (n_2) at (image3.south west) {(c)};
		\end{tikzpicture}
	\caption{Comparison of multi-view triangulation and local depth for estimating person keypoints in 3D: (a) multi-view triangulated 3D skeleton, (b) local depth from front view, and (c) local depth from side view. The local depth estimate results in a good approximation of the 3D skeleton for a front view, but is inaccurate in case of self-occlusion, e.g. from a side view. Multi-view triangulation is more robust but requires synchronization with other sensors.}
	\label{fig:local_depth}
	\vspace{-1.5em}
\end{figure}

The benefits of the occlusion information for the local sensor model are illustrated in \reffig{fig:occlusion}. Without occlusion feedback, heavy occlusion causes the pose estimation to collapse to the visible side only (\reffig{fig:occlusion}~(b)), which cannot be recovered by the feedback on the heatmap level~\cite{Bultmann_RSS_2021}. With occlusion information (\reffig{fig:occlusion}~(a)), unreliable, occluded joint detections can be discarded, and the local model is completed by the more reliable semantic feedback. Completely occluded persons can also be added back into the local model (\reffig{fig:occlusion}~(d)), making the sensor aware of persons that are going to re-appear in the future. Furthermore, the known occluded joints are excluded from multi-view triangulation in the next forward pass, as no new information can be gained from the respective sensor view.
In \refsec{sec:eval_quant}, we show that the added occlusion information improves the overall consistency in terms of reprojection error.

We further investigate the reliability of the local depth estimate of skeleton joints from the Jetson NX smart edge sensors.
The local depth enables estimating 3D joint positions from a single camera only, without dependence on other sensors. However, the RGB-D depth suffers from significant noise at larger distances and the depth measurement for a joint often is obstructed by occlusions or self-occlusions, as illustrated in \reffig{fig:local_depth}.
The local depth estimate results in a good approximation of the 3D skeleton for a front view, but is inaccurate in case of self-occlusion, e.g. from a side view.
Multi-view triangulation is more robust to these issues but requires synchronization with other sensors.

The local depth estimate, however, can still be used as an indication to constrain the data association between cameras for multi-view triangulation.
Person detections from different camera views are associated based on the epipolar distance of their joints using the efficient iterative greedy matching proposed by Tanke et al.~\cite{tanke_iterative_2019}. Keypoint detections from one image are projected as epipolar lines into the other cameras, where the distance from corresponding joint detections to the epipolar line is used as data-association cost. When a depth estimate is available, including an uncertainty interval computed from the keypoint confidence and the distribution of local depth readings, the matching can be restricted to a line segment. This helps to resolve ambiguous situations, where keypoints from multiple persons have a low distance to the epipolar line but are located at different positions along the line. Keypoints located on the line segment close to the projected depth estimate will receive lower data association cost while correspondences outside the projected depth interval will be discarded.

\section{Evaluation} %
\label{sec:Evaluation}
We evaluate the proposed system in challenging, cluttered real-world indoor scenes with multiple persons.
\subsection{Implementation Details}
Our sensor network consists of 20 smart edge sensors, thereof 4 based on the Jetson~NX board, as introduced in this paper, and 16 based on the Google Edge~TPU~\cite{Bultmann_RSS_2021}. The boards are connected to mains power supply and the power consumption of an NX board is \SIrange{20}{25}{\watt} during inference, thereof \SI{\sim 5}{\watt} for powering the RGB-D camera, compared to \SI{7}{\watt} for the Edge~TPU board.
The sensors cover an area of roughly 12$\times$22~m. The cameras face downward towards the center and run at \SI{30}{\hertz}.
We conduct experiments with the proposed, extended sensor network, with 8 persons moving in the covered area, which are evaluated in the following using a sequence of \SI{106}{\second} containing $\sim$\num[group-separator = {,}, group-minimum-digits = 4]{3000} frames per camera.

\subsection{Quantitative Results}
\label{sec:eval_quant}
\begin{table}[b]
\vspace{-1em}
\caption{Evaluation in real-world multi-person scenes with 20 cameras and 8 persons: Reprojection error (px) per joint class between detected 2D poses and fused 3D poses.}
\label{tab:error_2d_reproj}
\centering
\renewcommand{\arraystretch}{1.25} %
\linespread{0.95}\selectfont
\setlength{\tabcolsep}{3pt}
\begin{threeparttable}
\begin{tabular}{L{2cm}cc|cccccccc}
  \toprule
   Feedback & Cams & Pers & Head & Hips & Knees & Ankls & Shlds & Elbs & Wrists & \textbf{Avg}\\
  \midrule
   w/o fb   & 20  & 8 						 & 5.09 & 5.98 & 5.75 & 6.87 & 4.67 & 5.53 & 6.95 & 5.69 \\
   fb~\cite{Bultmann_RSS_2021} & 20 & 8 & 4.76 & 5.51 & 4.98 & 5.94 & 4.34 & 4.88 & 5.66 & 5.08 \\ %
   fb + occl. & 20 & 8 				 & 4.36 & 4.68 & 4.37 & 5.44 & 3.97 & 4.38 & \textbf{5.04} & 4.56 \\
   fb + occl. + local depth & 20 & 8 & \textbf{4.30} & \textbf{4.63} & \textbf{4.32} & \textbf{5.42} & \textbf{3.91} & \textbf{4.33} & \textbf{5.04} & \textbf{4.51} \\
   \bottomrule
\end{tabular}
\end{threeparttable}
\end{table}
To analyze the consistency between local and globally-fused human pose estimation, we evaluate the error between 2D poses detected in the individual sensor views and reprojected fused 3D poses in \reftab{tab:error_2d_reproj}. The reprojection error decreases for all joint classes when using the semantic feedback proposed in~\cite{Bultmann_RSS_2021} over a purely feed-forward pipeline. Adding the occlusion information, as proposed in this work, further decreases the reprojection error, as unreliable occluded keypoint detections can be discarded and excluded from multi-view triangulation. Constraining the data association using the local depth estimates of the RGB-D cameras gives a further small improvement.
The proposed pipeline leads to the lowest reprojection error for all joint classes, amounting to \SI{4.51}{\pixel} on average, indicating that the consistency between local and globally-fused pose estimation increases through the semantic feedback with occlusion information and by using local depth estimates in the data association step for multi-view triangulation.

\subsection{Qualitative Results}
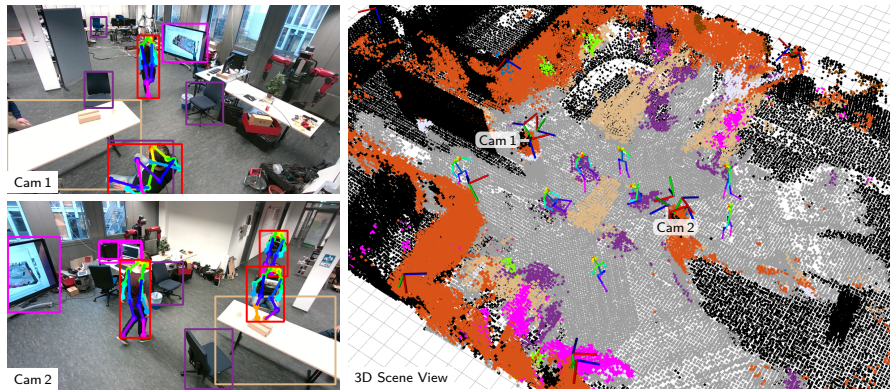
\begin{figure}[t]
	\centering
		\begin{tikzpicture}
			\node[inner sep=0,anchor=north west] (image1) at (0, 0) {\includegraphics[height=2.5cm]{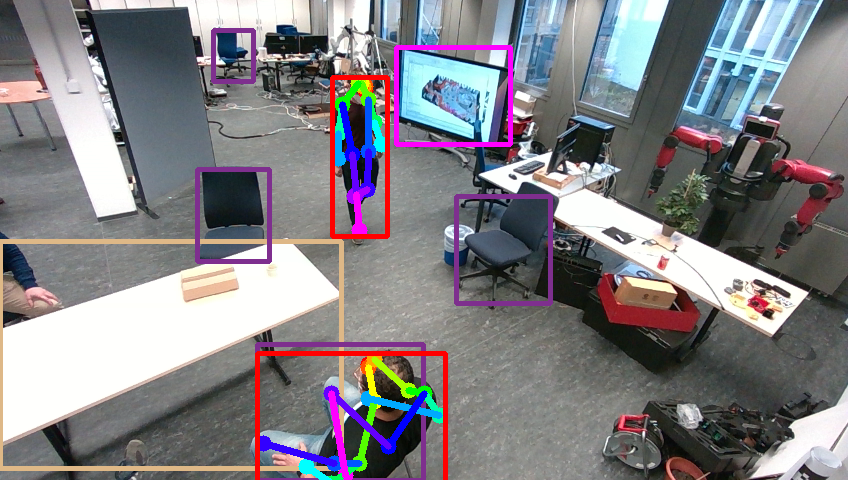}};
			\node[inner sep=0,anchor=north west,yshift=-0.1cm] (image2) at (image1.south west) {\includegraphics[height=2.5cm]{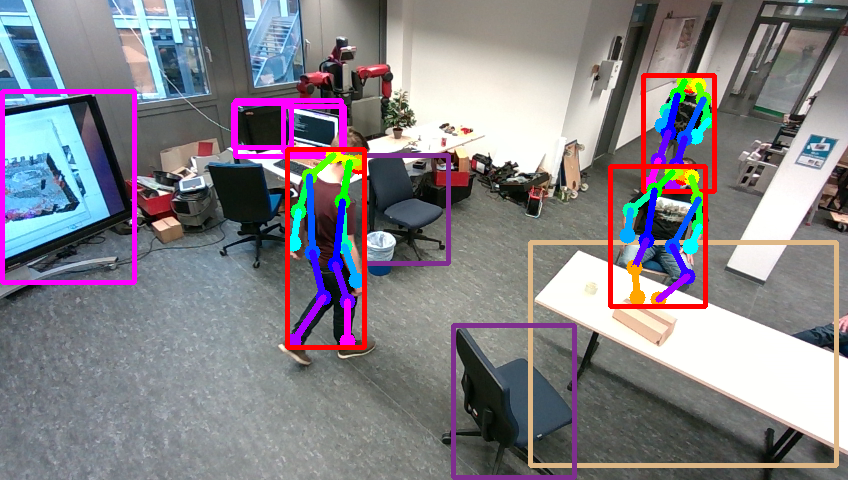}};
			\node[inner sep=0,anchor=north west,xshift=0.1cm] (image3) at (image1.north east) {\includegraphics[height=5.1cm,trim= 140 180 250 130, clip]{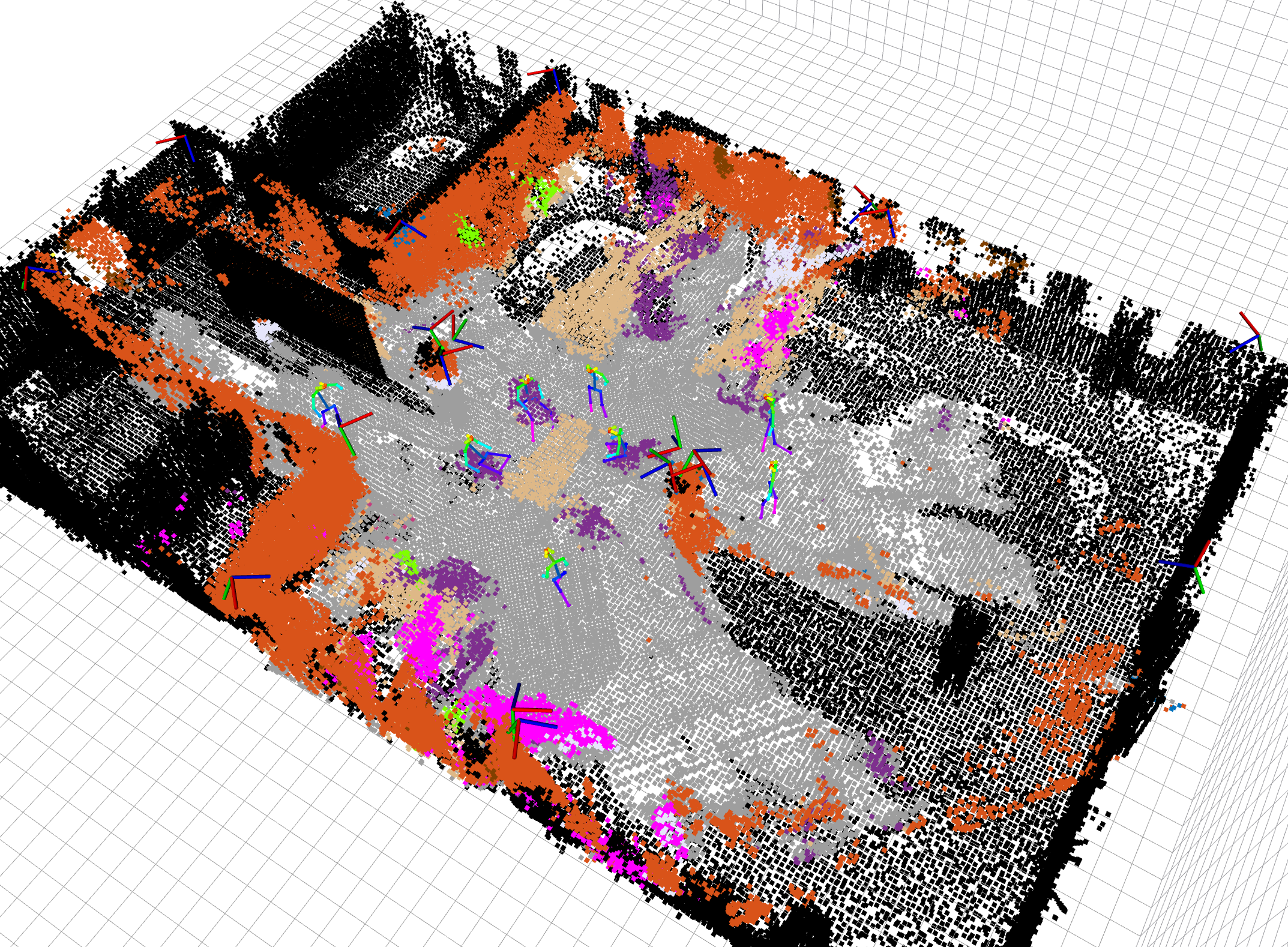}};
			
			\node[label,scale=.75, anchor=south west, rectangle, fill=white, align=center, font=\scriptsize\sffamily] (n_0) at (image1.south west) {Cam\,1};
			\node[label,scale=.75, anchor=south west, rectangle, fill=white, align=center, font=\scriptsize\sffamily] (n_1) at (image2.south west) {Cam\,2};
			\node[label,scale=.75, anchor=south west, rectangle, fill=white, align=center, font=\scriptsize\sffamily] (n_2) at (image3.south west) {3D Scene View};
			\node[label, scale=.75, anchor=south west, xshift=2.25cm, yshift=-2.55cm, rectangle, rounded corners=2, inner sep=0.06cm, fill=white,opacity=.8,text opacity=1, align=center, font=\scriptsize\sffamily] (l_cam1) at (image3.north west) {Cam\,1};
			\draw[white,opacity=.8, very thick] (l_cam1.358) ++(-0.005, 0) -- ++(0.12, 0);
    \node[label, scale=.75, anchor=south west, xshift=5.4cm, yshift=-4.1cm, rectangle, rounded corners=2, inner sep=0.06cm, fill=white,opacity=.8,text opacity=1, align=center, font=\scriptsize\sffamily] (l_cam2) at (image3.north west) {Cam\,2};
    \draw[white,opacity=.8, very thick] (l_cam2.135) ++(0, -0.005) -- ++(0, 0.16);
		\end{tikzpicture}
	\caption{Experiments in real-world multi-person scenes: Local detections of two reference sensor views and 3D semantic scene view with 3D poses of eight persons estimated in real-time. The geometry of the room is accurately represented by the map, fusing prior map (black) and current observations of smart edge sensors with their semantic classes (e.g., tables, chairs, computers). Interactions between persons and the scene, e.g., persons sitting on chairs (violet), are explained in a physically plausible way by the scene model.}
	\label{fig:scene_view}
\end{figure}

An exemplary scene of the real-world multi-person experiments is shown in \reffig{fig:scene_view}. Local detections and pose estimation in two reference camera views are depicted together with the 3D semantic scene view. 3D poses of eight persons are estimated online, in real-time during the experiment.
The semantic map represents the 3D geometry of the scene, fusing prior map and current sensor observations, including semantic class probabilities.
Interactions between persons and objects in the scene, e.g., persons sitting on chairs, are explained in a physically plausible manner by the scene model.

\subsection{Run-time Analysis}
\label{sec:runtime}
We analyze the run-time of CNN inference and the validation score on the respective training dataset (\cf\refsec{sec:semantic_perception}) on different embedded hardware accelerators and for different numerical precision for the employed models in \reftab{tab:model_runtime}. Thermal detector and RGB segmentation are only executed on the Jetson NX, as the Edge TPU does not have enough computational power to run all models in parallel. The run-times on Jetson NX in 16-bit floating-point mode (fp16) are comparable to 8-bit quantized (int8) inference on the Edge TPU.
The inference times roughly halve when using int8 precision on Jetson NX for the detectors and also decrease for the segmentation.
For pose estimation, here stated for a single crop and batch size 1, the difference is less significant. The inference time is only about \SI{4}{\milli\second}, and the precision is less relevant compared to other overhead from the inference framework in this case.
The validation score is given as bounding-box or keypoint mean average precision (mAP) for the \textit{person} class as defined for the COCO dataset~\cite{lin_coco_2014} for the detectors or pose estimation, respectively, and as mean intersection over union (mIOU) for the semantic segmentation. It decreases between \num{0.2} and \SI{1.2}{\percent} when using int8 precision instead of fp16. The slightly better performance of the RGB detector on the Edge~TPU can be explained as it was trained for the \textit{person} class only, while the detector used on Jetson~NX was trained for \textit{person} and 12 indoor object classes.
\begin{table}[t]
\caption{Average inference time and validation score (given as mAP for detectors and pose estimation and mIOU for segmentation) of CNN models (batch size~1) on different embedded hardware and for different numerical precision.} 
\label{tab:model_runtime}
\centering
\setlength{\tabcolsep}{2.8pt}
\begin{threeparttable}
\begin{tabular}{l|c|c|c|c|c|c|c}
  \toprule %
  Model & Input Res. & \multicolumn{2}{c|}{Edge TPU (int8)} & \multicolumn{2}{c|}{Jetson NX (fp16)} & \multicolumn{2}{c}{Jetson NX (int8)}\\
  & & time & val. score & time & val. score & time & val. score\\
  \midrule %
  RGB det. & 848$\,\times\,$480 & - & - & \SI{24.1}{\milli\second} & \SI{36.2}{\percent} & \textbf{\SI{11.8}{\milli\second}} & \SI{36.0}{\percent}\\
  RGB det. & 640$\,\times\,$480 & \SI{21.5}{\milli\second} & \textbf{\SI{36.7}{\percent}} & - & - & - & -\\
  Pose est. & 192$\,\times\,$256 & \SI{4.5}{\milli\second} & \SI{68.4}{\percent} & \SI{4.0}{\milli\second} & \textbf{\SI{69.3}{\percent}} & \textbf{\SI{3.5}{\milli\second}} &  \SI{68.6}{\percent}\\
  Thermal det. & 160$\,\times\,$120 & - & - & \SI{13.9}{\milli\second} & \textbf{\SI{25.4}{\percent}} & \textbf{\SI{6.0}{\milli\second}} & \SI{24.9}{\percent} \\
  RGB segm. & 849$\,\times\,$481& - & - & \SI{27.0}{\milli\second} &  \textbf{\SI{50.0}{\percent}} & \textbf{\SI{20.0}{\milli\second}} &  \SI{48.8}{\percent}\\
  \bottomrule %
\end{tabular}
\end{threeparttable}
\end{table}

\begin{table}[t]
\caption{Average processing time for pose estimation (inference + post-processing) for increasing number of person detections per image. Batch processing can be used on Jetson NX while crops are processed one by one on the Edge TPU.} 
\label{tab:pose_runtime}
\centering
\setlength{\tabcolsep}{3pt}
\begin{threeparttable}
\begin{tabular}{lc|cccccc}
  \toprule %
  Sensor Type & Precision & 1 & 2 & 3 & 4 & 5 & 6\\
  \midrule %
  Edge TPU & int8 & \SI{14.4}{\milli\second} & \SI{23.6}{\milli\second} & \SI{33.0}{\milli\second} & \SI{43.9}{\milli\second} & \SI{54.3}{\milli\second}& \SI{65.9}{\milli\second}\\
  Jetson NX & int8 & \SI{12.1}{\milli\second} & \textbf{\SI{15.0}{\milli\second}} & \textbf{\SI{18.0}{\milli\second}} & \SI{27.4}{\milli\second} & \textbf{\SI{31.0}{\milli\second}} & \textbf{\SI{38.8}{\milli\second}}\\
  Jetson NX & fp16 & \textbf{\SI{11.8}{\milli\second}} & \SI{17.4}{\milli\second} & \SI{21.9}{\milli\second} & \textbf{\SI{26.4}{\milli\second}} & \SI{31.7}{\milli\second}& \SI{41.5}{\milli\second}\\
  \bottomrule %
\end{tabular}
\end{threeparttable}
\end{table}
Table~\ref{tab:pose_runtime} shows the scaling of processing time for pose estimation, including CNN inference and post-processing, with an increasing number of person detections per image. During our experiments with 8 persons in the scene, a maximum of 6 persons were visible at a time in one camera.
On the Edge TPU sensors~\cite{Bultmann_RSS_2021}, crops are processed one by one, as only a batch size of one is supported, and the runtime scales linearly. Up to three persons can be tracked at the full camera frame rate of \SI{30}{\hertz}.
On the Jetson NX platform, batch-processing is possible, and therefore the run-times scale sub-linearly. Up to five persons can be tracked at the full camera frame rate.
For pose inference, there is only a small difference in run-time between fp16 and int8 mode on Jetson~NX.

We run CNN inference in fp16 mode on the Jetson~NX smart edge sensors during our online experiments to benefit from the higher numerical precision, as 8-bit quantization gives only little gains in run-time for the pose estimation with strong real-time constraints, and the fp16 inference time is sufficient for the other models that run with lower priority at a \SI{1}{\hertz} update rate.

\section{Conclusion}
\label{sec:Conclusion}
In this work, we presented a network of distributed smart edge sensors for 3D semantic scene perception, including static or slowly moving geometry as well as dynamic human motions.
RGB-D and thermal camera images are processed locally on the sensor boards with vision CNNs for person and object detection, semantic segmentation, and human pose estimation. 2D human keypoint estimations, augmented with the RGB-D depth estimate, as well as semantically annotated point clouds are streamed from the sensors to a central backend, where multiple viewpoints are fused into an allocentric 3D semantic scene model.
The individual sensors incorporate global context information into their local models via a semantic feedback channel. For this, the globally-fused 3D human poses are projected into the sensor views, where they are fused with the local detections. The estimated 3D geometry enables to add occlusion information for each joint to the semantic feedback, such that unreliable, occluded joint detections can be discarded on the sensors, and the local models can be complemented by the more reliable feedback joint positions.
We built a sensor network of 20 smart edge sensors, thereof 4 based on the novel Jetson NX board, covering an area of about 12$\times$22~m, and evaluated the proposed system in challenging, cluttered real-world scenes with up to 8 persons.
Dynamic human motions are estimated in real time and the semantically annotated 3D geometry provides a complete scene view that also explains interactions between persons and objects in the scene.

Future work includes using the 3D semantic scene model and human poses estimated by the smart edge sensors to enable anticipative robot behavior and safe human-robot interaction in a shared workspace. Mobile sensor nodes could further be added to the sensor network for active exploration of areas not covered by the permanently installed sensors.

\section*{Acknowledgments}
This work was funded by grant BE 2556/16-2 of the German Research Foundation (DFG) and Fraunhofer IAIS.

\bibliographystyle{splncs03}
\bibliography{literature}

\end{document}